\def\year{2022}\relax
\documentclass[letterpaper]{article} 
\usepackage{aaai22}  
\usepackage{times}  
\usepackage{helvet}  
\usepackage{courier}  
\usepackage[hyphens]{url}  
\usepackage{graphicx} 
\usepackage{natbib}  
\usepackage{caption} 
\DeclareCaptionStyle{ruled}{labelfont=normalfont,labelsep=colon,strut=off} 
\usepackage{amsmath, amssymb,amsfonts}
\usepackage{textcomp}
\usepackage{comment}
\usepackage{subcaption}
\usepackage{multirow}
\usepackage{tabularx}
\usepackage{algcompatible}
\usepackage{relsize}
\usepackage{stmaryrd}
\usepackage{graphicx,dblfloatfix}
\usepackage{bbm}
\usepackage[ruled]{algorithm2e}
\usepackage{algorithmicx}
\usepackage{algpseudocode}

\title{TrustAL: Trustworthy Active Learning using Knowledge Distillation}
\author {
    Beong-woo Kwak\textsuperscript{\rm 1},~
    Youngwook Kim\textsuperscript{\rm 2},~
    Yu Jin Kim\textsuperscript{\rm 1},~
    Seung-won Hwang\textsuperscript{\rm 3},~
    Jinyoung Yeo\textsuperscript{\rm 1}\thanks{Corresponding author}
}
\affiliations {
    \textsuperscript{\rm 1} Department of Artificial Intelligence, Yonsei University\\
    \textsuperscript{\rm 2} Department of Computer Science, Yonsei University\\
    \textsuperscript{\rm 3} Department of Computer Science and Engineering, Seoul National University\\
    \texttt{\{beongwoo.kwak,youngwook,yujin000731\}@yonsei.ac.kr\\seungwonh@snu.ac.kr\\jinyeo@yonsei.ac.kr}
}

\usepackage{bibentry}

\newcommand{\eg}{{\it e.g.}}%
\newcommand{\ie}{{\it i.e.}}%
\newcommand{\calA}{\mbox{${\cal A}$}}

\newcommand{\calD}{\mbox{${\cal D}$}}

\newcommand{\calL}{\mbox{${\cal L}$}}

\newcommand{\calQ}{\mbox{${\cal Q}$}}

\newcommand{\calU}{\mbox{${\cal U}$}}

\begin{document}
\maketitle

\begin{abstract}
Active learning can be defined as iterations of data labeling, model training, and data acquisition, until sufficient labels are acquired. A traditional view of data acquisition is that, through iterations, knowledge from human labels and models is implicitly distilled to monotonically increase the accuracy and label consistency. Under this assumption, the most recently trained model is a good surrogate for the current labeled data, from which data acquisition is requested based on uncertainty/diversity. Our contribution is debunking this myth and proposing a new objective for distillation. First, we found example forgetting, which indicates the loss of knowledge learned across iterations. Second, for this reason, the last model is no longer the best teacher-- For mitigating such forgotten knowledge, we select one of its predecessor models as a teacher, by our proposed notion of ``consistency''. We show that this novel distillation is distinctive in the following three aspects; First, consistency ensures to avoid forgetting labels. Second, consistency improves both uncertainty/diversity of labeled data. Lastly, consistency redeems defective labels produced by human annotators.
\end{abstract}

\section{Introduction} \label{sec:intro}

Labeling data is a fundamental bottleneck in machine learning due to annotation cost and time. One practical solution is Active Learning (AL): given a limited labeling budget $k$, which example should I ask human annotators to label?  Generally, this can be done through an \emph{iterative} process of labeling data, model training, and data acquisition steps, until sufficient labels are obtained. At each iteration, based on the last trained model, unlabeled yet the $k$ most desirable examples are recognized and added to the labeled dataset to train a new model. This process continues to the next iteration for selecting next $k$ unlabeled examples based on the newly trained model. That is, a naive belief in AL is that the last trained model can be a good reference or surrogate for the distribution of the currently labeled data, which indicates what examples are desired for the next model update.

In this work, our empirical observation dispels this myth. Although the model knowledge learned from the labels is expected to be ``consistently'' kept or improved across AL iterations, we find that knowledge learned at some time is suddenly forgotten, which indicates that the recent model is ineligible to be treated as a good reference of the labeled dataset. More specifically, we can observe such inconsistent behaviors of the trained model during inference time, where sample $i$ predicted correctly at iteration $t$ is predicted incorrectly at iteration $t + \Delta t$, which is called \emph{example forgetting}~\cite{toneva2018empirical}.

Motivated by this, in this work, we argue that \textbf{correct-consistency} (which we call consistency for brevity) should be an essential criterion, which is the model ability to make consistent correct predictions across successive AL generations for the same input~\cite{wang2020wisdom}. In the view of consistency, prior AL methods only focusing on data acquisition steps~\cite{dasgupta2011two,xu2003representative,bodo2011active,ash2019deep} are still sub-optimal since the three transitions among AL steps may suffer from following problems due to inconsistency (reverse phenomenon of consistency),  which we empirically analyze later:
\begin{itemize}
    \item \textbf{From labeling to model training:} Despite successful data acquisition, the subsequent labeling efforts can be negated by forgetting the learned knowledge later, which wastes annotation cost. We argue that consistency is key to make a label-efficient AL (Figure~\ref{fig:rq1-Trec}). 

    \item \textbf{From model training to data acquisition:} Inconsistent data acquisition models cannot serve as a good reference for the current data distribution, which leads to contaminating the next data acquisition step. Improving consistency may be synergetic to either uncertainty- or diversity-based acquisition strategies (Figure~\ref{fig:rq2}).

    \item \textbf{From data acquisition to labeling:} Human annotators who act as oracles are usually subject to accidental mislabeling~\cite{bouguelia2018agreeing} which degrades traditional AL methods. Learning to keep consistency enables to mitigate the confusion from the noisy labels (Figure~\ref{fig:noisy_label}).
\end{itemize}

To overcome these drawbacks and thus make all the three transitions in AL more \textbf{trustworthy}, we propose a label-efficient AL framework, called \textbf{Trust}worthy \textbf{AL} (\textbf{TrustAL}), for bridging the knowledge discrepancy between labeled data and model. In TrustAL, our key idea is to add a new step in the iterative process of AL to learn the forgotten knowledge, which is orthogonally applicable to state-of-the-art data acquisition strategies in a synergistic manner. Specifically, at each iteration, TrustAL first searches for an expert model for the forgotten knowledge among the predecessor models. Then, TrustAL transfers the model knowledge (\eg, logits) to the current model training step by leveraging the knowledge distillation technique~\cite{hinton2015distilling}. By optimizing the dual goals of following both human labels and machine labels of the expert predecessor, the newly trained model can relieve forgotten knowledge and thus be more consistent, keeping its correct predictions.

For the purpose of identifying which predecessor is the most desired teacher to relieve the forgotten knowledge, we further explore the teacher selection problem. To resolve this, we present two teacher selection strategies, (1) \textbf{TrustAL-MC}: monotonic choice of the most recent model (\ie, a proxy of the most accurate model), 
and (2) \textbf{TrustAL-NC}: non-monotonic choice of the well-balanced model with accuracy and consistency, which we thoroughly design as analysis/evaluation measures in this paper.

Our experiments show that the TrustAL framework significantly improves performance with various data acquisition strategies while preserving the valuable knowledge from the labeled dataset. We validate the pseudo labels from the predecessor models are not just approximate/weak predictions \text{-} It can be viewed as knowledge from the previous generation, and can be used as consistency regularization for conventional AL methods solely aiming at higher accuracy.

\section{Preliminaries \& Related Work} \label{sec:pre}

\subsection{Active Learning for Classification}\label{sec:pre.1}

Given an arbitrary classification task, assume that there is a large unlabeled dataset $\calU = \{x_i\}_{i=1}^n$ of $n$ data samples. The goal of AL is to sample a subset $\calQ \subset \calU$ to efficiently label so that newly training a deep neural network parameters $\theta$ for the classifier $f(x;\theta)$ improves test accuracy. Algorithm~\ref{alg:al} describes the conventional procedure in AL. On each iteration $t$, the learner uses strategy $\calA$ (\eg, uncertainty or diversity) to acquire $k$ samples $\calQ_t$ from dataset $\calU$. Generally, data acquisition model $M_t$ is used for evaluating unlabeled samples according to $\calA$. Then, for sample $x_i$, the learner queries for its oracle label $y_i \in 1,...,c$, where $c$ is the number of classes. We denote the predicted label of trained model $\theta_t$ for $x_i$ by $\hat{y}_i^t = argmax_{c} f(y_{ic}|x_i;\theta_t)$. 

\begin{algorithm}[h!]
\fontsize{9.5}{11.5}\selectfont
\STATE{\textbf{Input:} Initial labeled data pool $\calL$, unlabeled data pool $\calU$, number of queries per iteration (budget) $k$, number of iterations $T$, sampling algorithm $\calA$}\\
\STATE{\textbf{Output:} Model parameters $\theta_T$} \\
\STATE{$\theta_0 \leftarrow$ Train a seed model on $\calL$}\\
\For{iteration $t=1,...,T$}{
    \STATE{$M_t(x) = f(x;\theta_{t-1})$}\\
    \STATE{$\calQ_t$ $\leftarrow$ Apply $\calA(x, M_t, k)$ for $\forall x \in \calU$}\\
    \STATE{$\bar{\calQ}_t \leftarrow $ Label queries $\calQ_t$ by oracles}\\
    \STATE{$\calL \leftarrow \calL \cup \bar{\calQ}_t$}\\
    \STATE{$\calU \leftarrow \calU \setminus \bar{\calQ}_t$}\\
    \STATE{$\theta_t \leftarrow$ Train a new model on $\calL$}\\
}
\STATE{\textbf{return}~ $\theta_T$}
\caption{Conventional AL procedure}\label{alg:al}
\end{algorithm}

In most AL approaches, a data acquisition model at time $t$ corresponds to the trained classification model at time $t\text{-}1$, \ie, $M_t = \theta_{t-1}$. We call this \textbf{monotonic acquisition}, since a naive belief would be assuming the last trained model $\theta_{t-1}$ is effective enough to not only provide a good representation for the entire labeled data $\calL$ but also estimate acquisition factors (\eg, confidence) for remaining unlabeled data $\calU$.

\subsection{Data Acquisition Strategies in AL}\label{sec:pre.2}
The ultimate goal of AL is to improve the classification accuracy with a fixed annotation budget~\cite{settles2009active,lowell2018practical}. Existing research efforts on pool based active learning~\cite{lewis1994sequential} achieve this goal by focusing on data acquisition based on query strategy and data strategy~\cite{ren2020survey}. As a query strategy, there are two general approaches to recognize the most appropriate samples~\cite{dasgupta2011two} with monotonic acquisition: uncertainty sampling and diversity sampling. While uncertainty sampling efficiently searches the hypothesis space by finding difficult examples to label~\cite{asghar2017deep,he2019towards,ranganathan2017deep}, diversity sampling exploits heterogeneity in the feature space~\cite{hu2010off,bodo2011active}. Recently, hybrid approaches are proposed \cite{zhdanov2019diverse,ash2019deep}. Particularly, BADGE~\cite{ash2019deep} successfully integrates both aspect by clustering hallucinated gradient vectors based on monotonic acquisition scheme. 


\subsection{Data Acquisition Models in AL}\label{sec:pre.3}
Despite the remarkable success in query strategies, recent research has concerned several limitation of AL. \citet{yun2020weight,wang2016cost} point out the difficulty of learning good representation across AL iterations since insufficient annotations may lead to the instability of training models. This indicates that the monotonic acquisition does not ensure the last trained model as a good surrogate of the currently labeled data to identify the informative samples for data acquisition. As a result, \citet{karamcheti-etal-2021-mind,farquhar2020statistical,prabhu2019sampling} reveal that the acquired samples are vulnerable to sampling bias, and especially, \citet{karamcheti-etal-2021-mind} have presented Dataset Maps~\cite{swayamdipta2020dataset} of AL, which visualizes harmful outliers preferred by AL methods. Despite these facts, \citet{lowell2018practical} suggest that the monotonic acquisition is still promising as another remedy using external models (\eg, SVM out of the AL iterations) for data acquisition extremely hampers accuracy of AL.

Motivated by this line of research, in this work, we explore how the limitation of monotonic acquisition can be addressed, in particular, considering consistency as a solution to mitigate the instability of AL iterations. Similar to~\cite{lowell2018practical} reporting unreliable performance of AL in the NLP field, we choose text classification tasks as our testbed. 

\begin{figure*}[t!]
     \centering
     \begin{subfigure}[b]{0.24\textwidth}
         \centering
         \includegraphics[width=\textwidth]{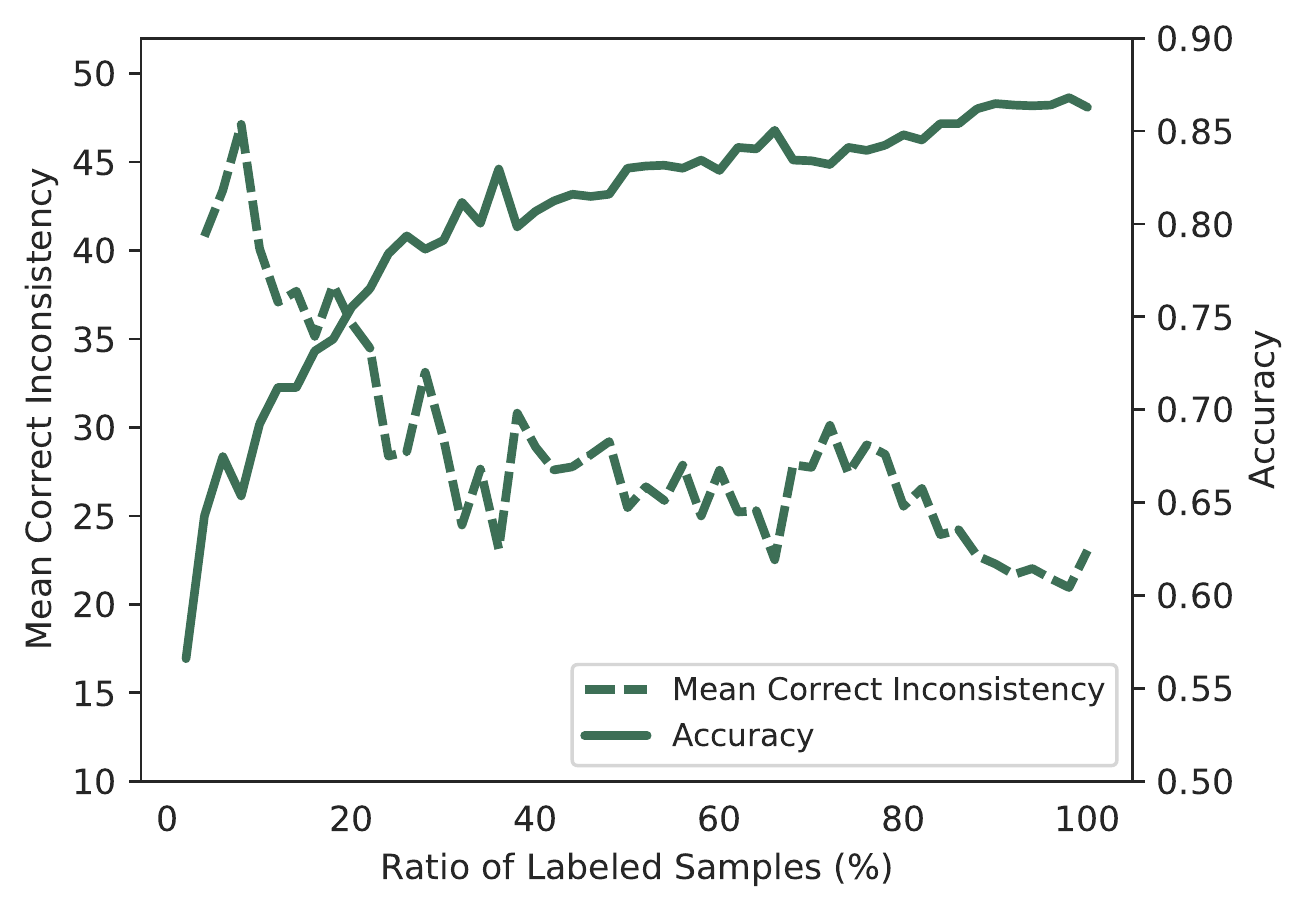}
         \caption{Random sampling}
     \end{subfigure}
     \hfill
     \begin{subfigure}[b]{0.24\textwidth}
         \centering
         \includegraphics[width=\textwidth]{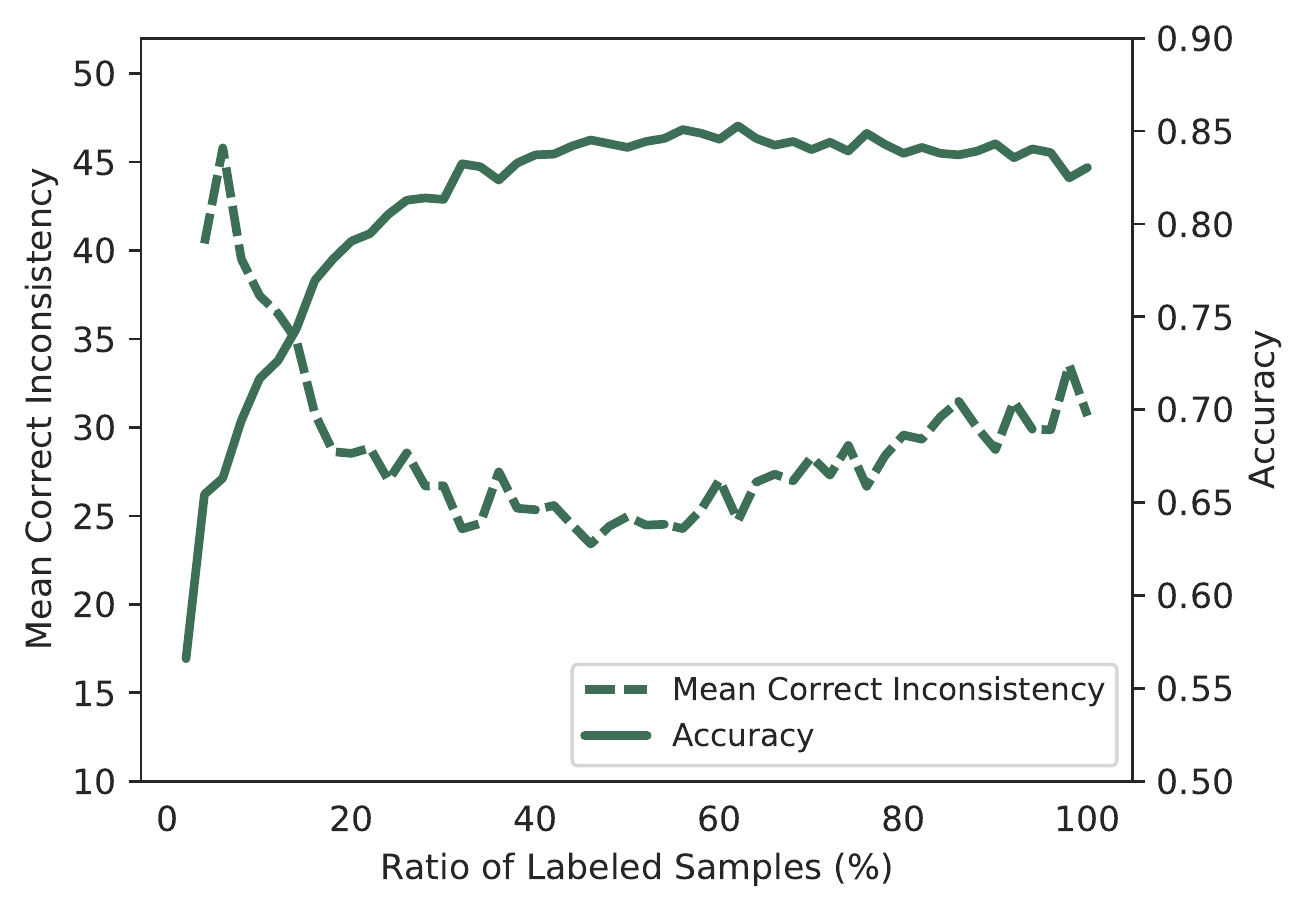}
         \caption{CONF}
     \end{subfigure}
     \hfill
     \begin{subfigure}[b]{0.24\textwidth}
         \centering
         \includegraphics[width=\textwidth]{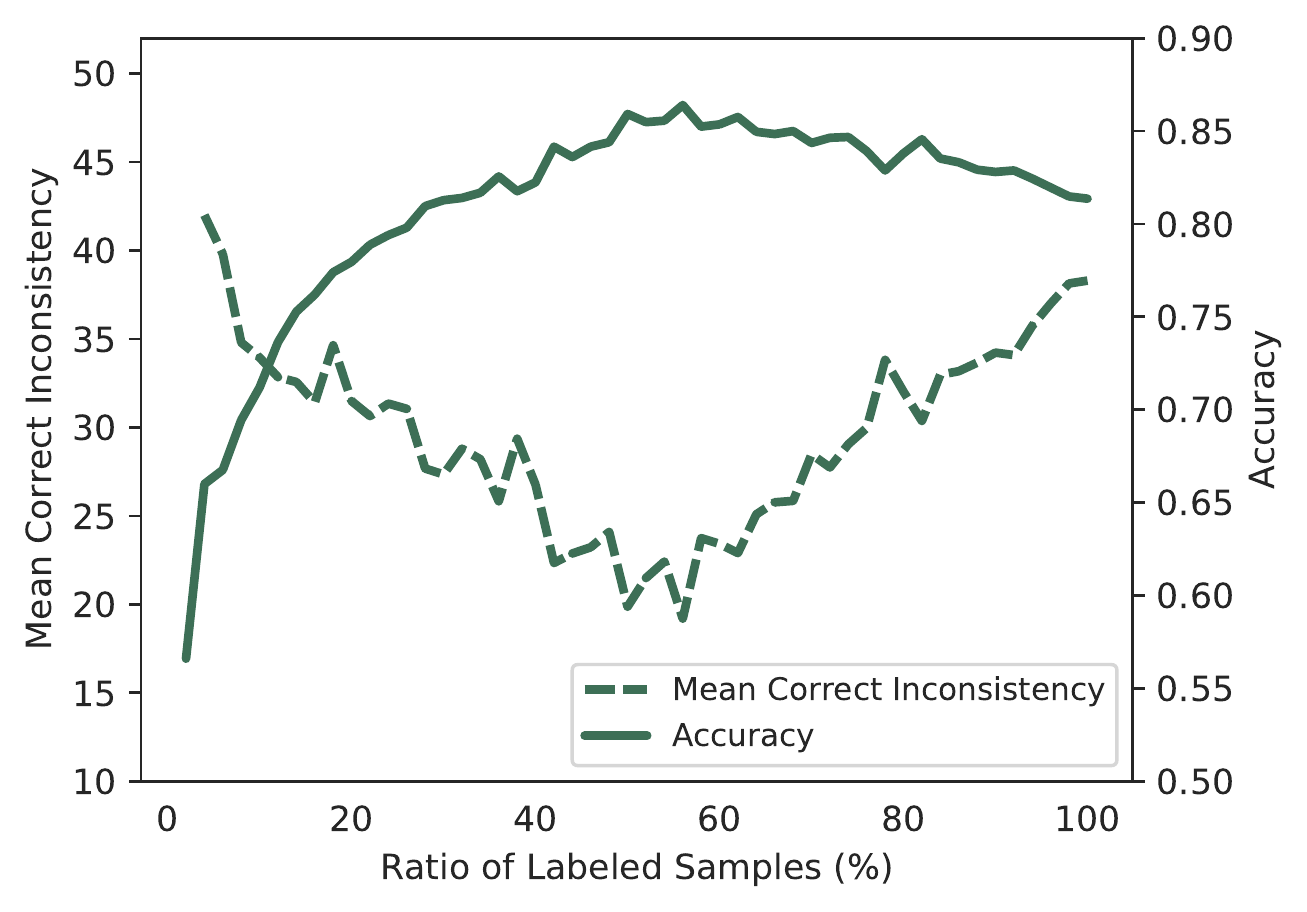}
         \caption{CORESET}
     \end{subfigure}
     \begin{subfigure}[b]{0.24\textwidth}
         \centering
         \includegraphics[width=\textwidth]{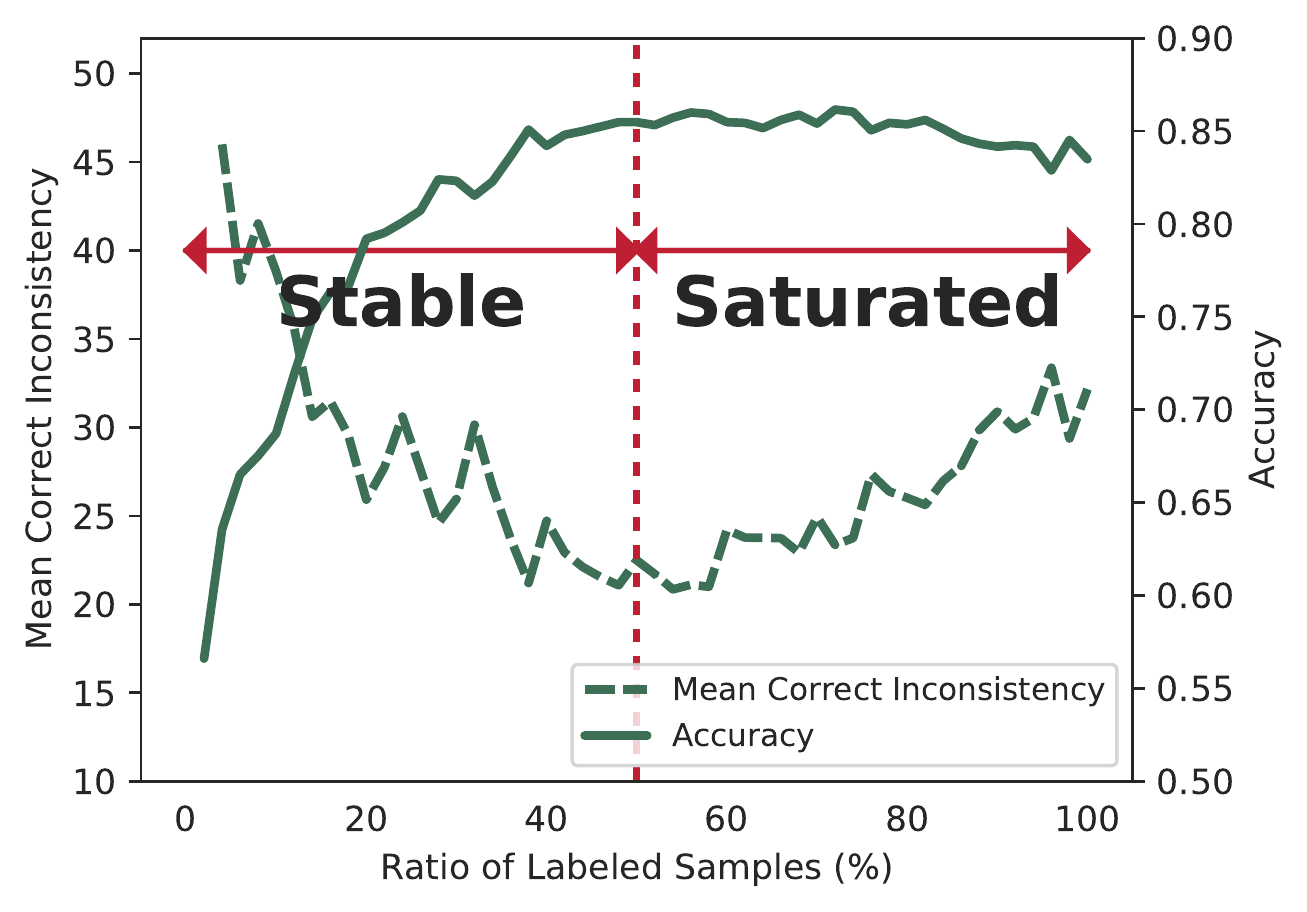}
         \caption{BADGE}
     \end{subfigure}
     \hfill
     \begin{subfigure}[b]{0.24\textwidth}
         \centering
         \includegraphics[width=\textwidth]{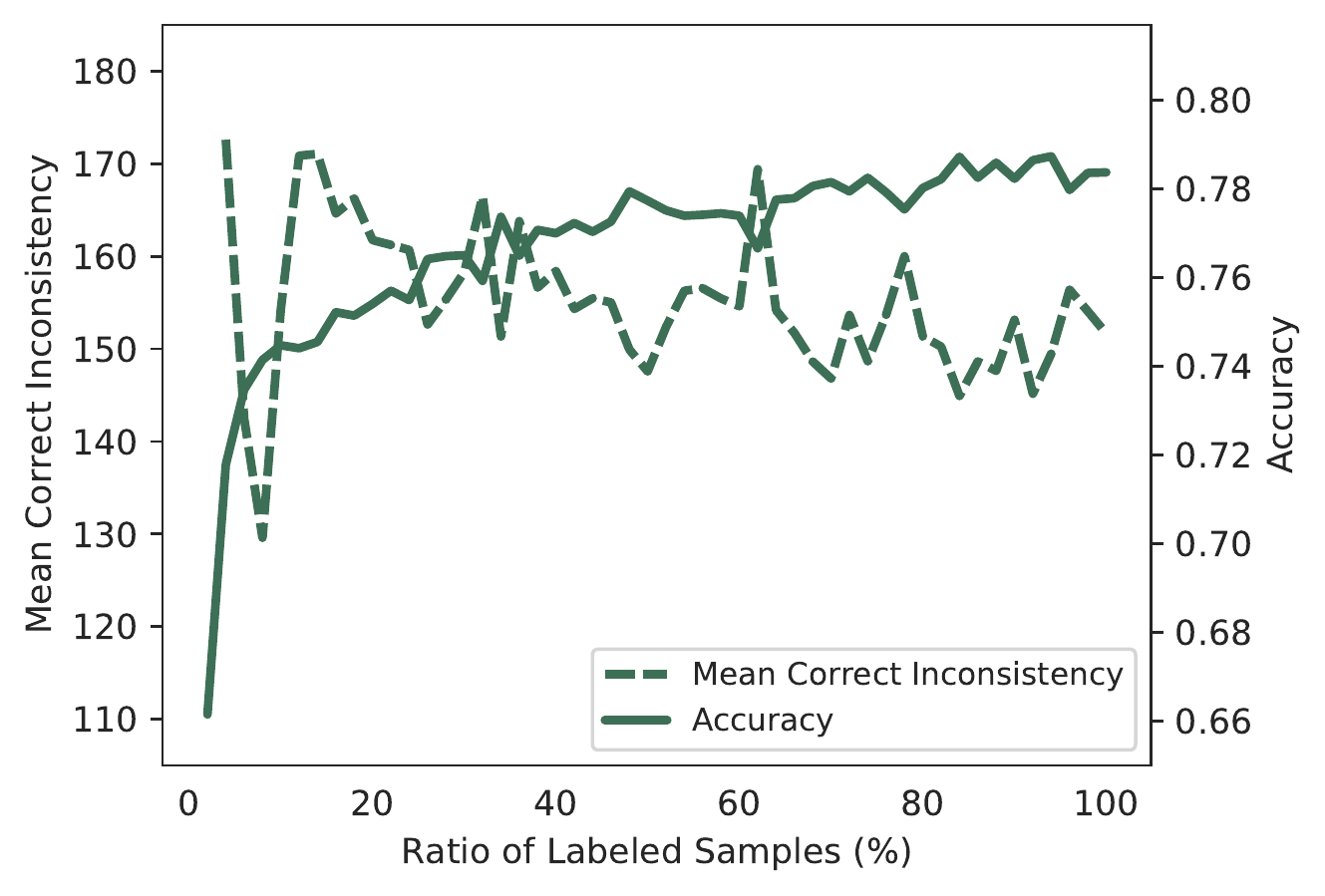}
         \caption{Random sampling}
     \end{subfigure}
     \hfill
     \begin{subfigure}[b]{0.24\textwidth}
         \centering
         \includegraphics[width=\textwidth]{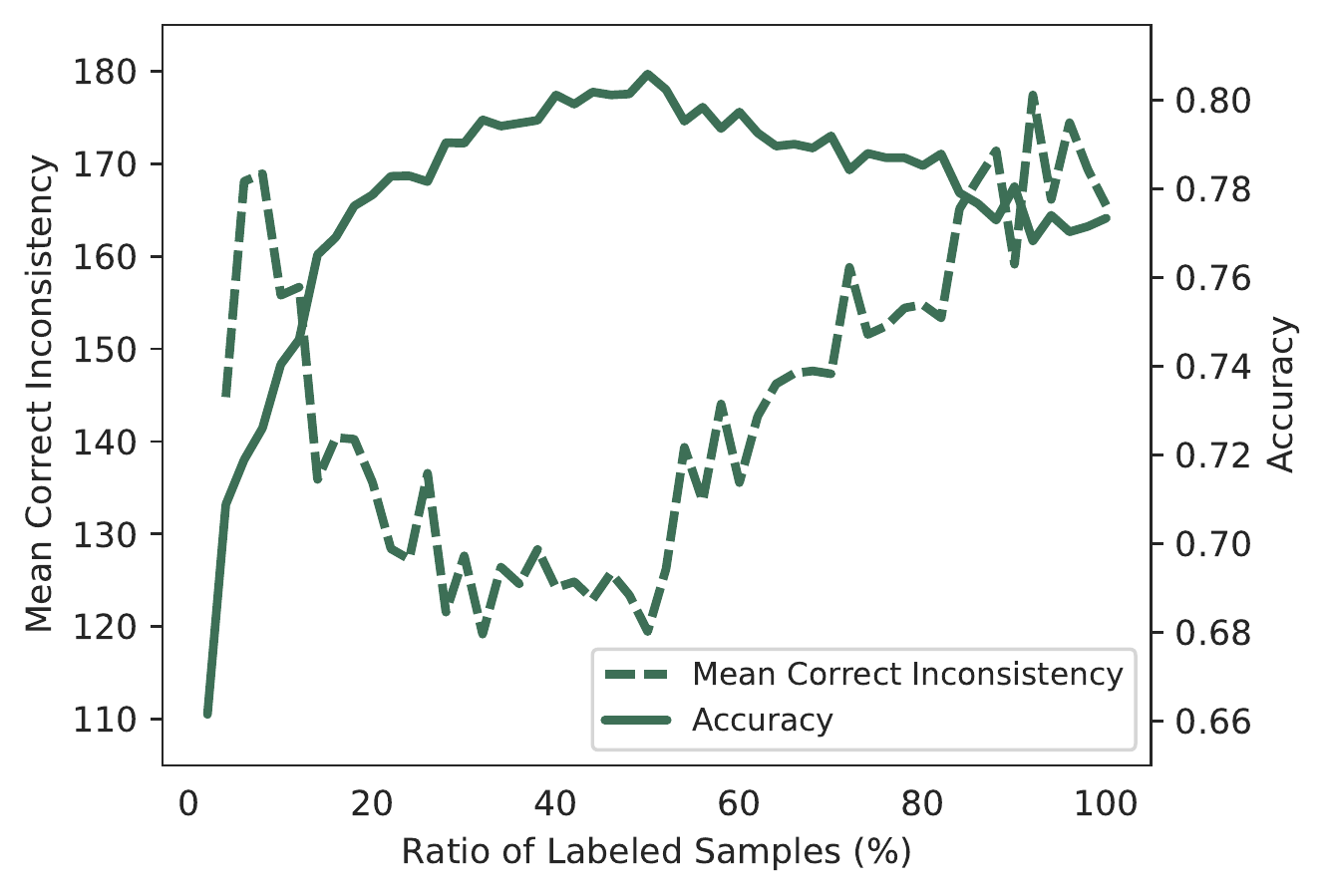}
         \caption{CONF}
     \end{subfigure}
     \hfill
     \begin{subfigure}[b]{0.24\textwidth}
         \centering
         \includegraphics[width=\textwidth]{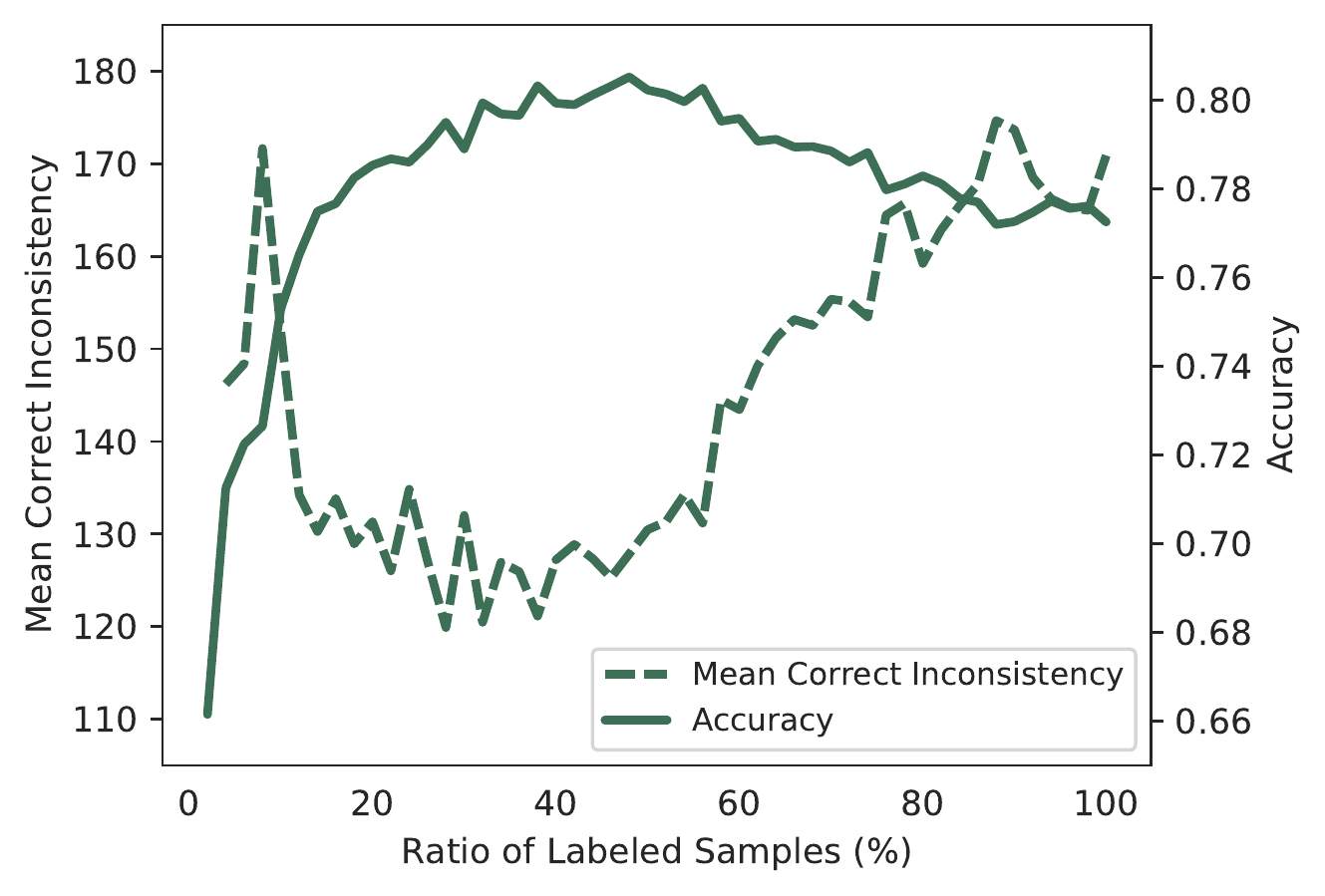}
         \caption{CORESET}
     \end{subfigure}
     \begin{subfigure}[b]{0.24\textwidth}
         \centering
         \includegraphics[width=\textwidth]{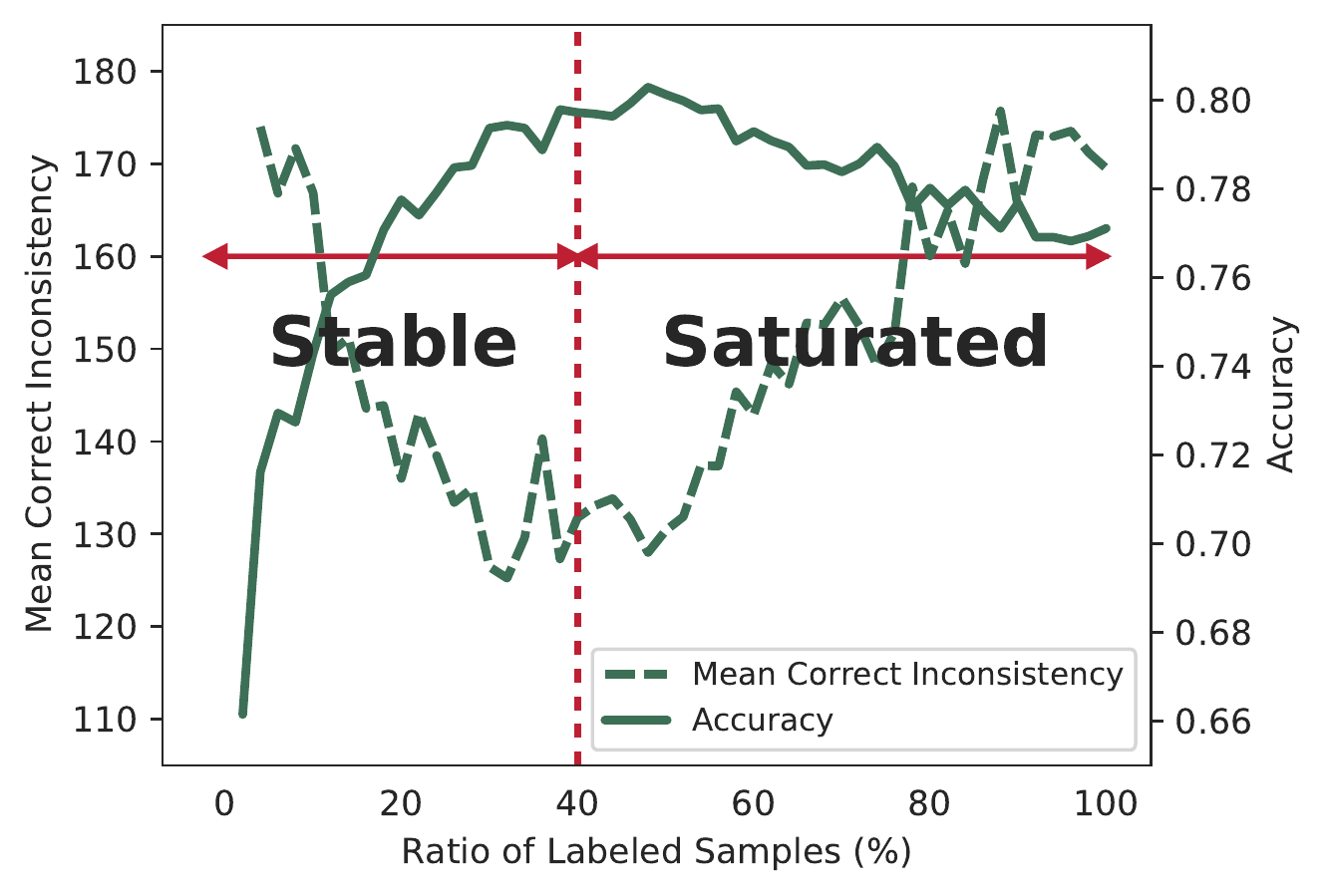}
         \caption{BADGE}
     \end{subfigure}
    \caption{The accuracy and consistency (MCI) of Bi-LSTM model under three acquisition strategies on two text classification test datasets (TREC (a\text{-}d) and SST-2 (e\text{-}h)); x-axis represents the ratio of labeled samples and y-axis represents the corresponding metrics; We report the average values with five random seeds.}
    \label{fig:exampleforgetting}
\end{figure*}

\section{Accuracy-Consistency Dynamics}\label{sec:analysis.2}
In this section, we analyze the training dynamics of AL in terms of consistency along with accuracy, observing (catastrophic) \emph{example forgetting event}~\cite{toneva2018empirical} on the AL iterations: these occur when examples that have been ``learnt'' (\ie, correctly classified) at some time $t$ in the optimization process are subsequently misclassified -- or in other terms ``forgotten'' -- at a time $t\text{+}\Delta t > t$. Formally,

\newtheorem{definition}{Definition}
\begin{definition}{(Forgetting and Learning Events)}\label{def:forgetting}
Given a predicted label $\hat{y}_i^t$, let $\text{acc}_i^t = \mathbbm{1}_{\hat{y}^t_i = y_i}$ be a binary variable indicating whether the example is correctly classified by $\theta_t$. Sample $i$ undergoes a forgetting event when $acc_i^t$ decreases between two different iterations, \ie, $acc^{t}_i > acc^{t + \Delta t}_i$, where $\Delta t > 0$ (misclassified). Conversely, a learning event has occurred if $acc^{t}_i < acc^{t + \Delta t}_i$. 
\end{definition}

While learning new knowledge is also one of the important factors for generalization ability, our focus is on measuring how well models in AL preserve the learned knowledge. 
For further analysis, we introduce \textit{Correct Inconsistency} of a model as a measure of how inconsistent the target model is with its predecessor models for a sample. That is, correct inconsistency counts the forgetting events between the model and each of the predecessor models.

\begin{definition}{(Correct Inconsistency)}\label{def:ci}
The degree of correct inconsistency of $\theta_t$ for sample $x_i$ is measured as the number of occurrences of \textit{forgetting events} for sample $x_i$ from any predecessor model $\theta_{t-\Delta t}$, where $0 < \Delta t \leq t$:
\begin{equation*}
    \mathbb{C}^{(t)}_i = \sum_{\Delta t=1}^{t} \mathbbm{1}_{(acc^{t - \Delta t}_i > acc^{t}_i)}
\end{equation*}
\end{definition}

As the number of predecessor models are different per AL iteration, to fairly show the degree of inconsistency, we use mean of correct inconsistency for every sample in development split \eg, MCI =$ \sum_{i}{\mathbb{C}^{(t)}_i/t}$.

In Figure~\ref{fig:exampleforgetting}, we present both accuracy and MCI of trained models through the full progress of AL. We analyze three data acquisition strategies that are carefully chosen by considering the uncertainty-diversity dichotomy~\cite{yuan2020cold} along with random strategy. CONF~\cite{wang2014new}, CORESET~\cite{sener2017active}, and BADGE~\cite{ash2019deep} represent three lines of acquisition strategies in AL: uncertainty, diversity, and their hybrid. Across all dataset and acquisition strategies, accuracy and MCI follow the anti-correlated relationship. For convenience of analysis, we divide the training progress into two phases based on the transition of tendency in terms of accuracy: \emph{stable} and \emph{saturated} phases.

In the stable phase, more data leads to more accurate model. Validation accuracy increases on 0-50\% of TREC and 0-40\% of SST-2, while MCI decreases, where newly labeled samples improve generalization of the trained models. In this phase, AL strategies seek to achieve label efficiency, reaching higher accuracy with a given amount of labeled samples or reversely achieving the same accuracy with less amount of labeled samples. What is notable here is that dramatic improvement of accuracy mostly involves the rapid drop in MCI. These analytical results provide a guide towards the idealistic property of AL methods, which is preserving existing knowledge and simultaneously learning new knowledge. Thus, we hypothesize that relieving forgetting events may contribute to faster and better (\ie, higher accuracy) convergence of AL.

In the saturated phase, the monotonic trends observed in stable phase do not hold. Validation accuracy converges or decreases with the rapid increase of MCI, suggesting that the generalization performance of model deteriorates as even more labeled samples are fed into the trained models. That is, more data does not always lead to more accurate model, which indicates labeling efforts may be negated in this phase. Although such an extremely undesirable situation in AL is barely addressed by stopping AL iterations in prior work~\cite{ishibashi2020stopping}, an idealistic AL framework would avoid this phase so that the models can be learned in a more label-efficient manner.

\section{TrustAL: Trustworthy Active Learning}\label{sec:trustal}
Based on the prior findings on training dynamics of AL procedure, we aim to relieve forgotten knowledge to train better acquisition model that serves as a good surrogate for labeled dataset. A naive way to obtain more generalized models is simply using enough labeled data. However, this approach is not always applicable since budget is limited in AL. Another line of work is using multiple equivalent models (\eg, ensemble) based on complementary nature across different generations. However, this approach is also not always affordable since querying on the huge pool of unlabeled data using multiple models is computationally too expensive.

We now present \textbf{TrustAL} (Trustworthy AL) that enables the training of consistent acquisition model that serves as a good reference for labeled dataset in smart and resource-efficient way. \textbf{TrustAL} utilizes additional machine generated labels for the purpose of mitigating the forgotten knowledge. Especially, among predecessor models, \textbf{TrustAL} identifies a proper expert model that can efficiently contribute to mitigating forgotten samples, which is a novel way of tackling the possible knowledge loss during AL procedure.

\subsection{Distillation-based Consistency Regularization}\label{sec:trustal.1}
Knowledge distillation~\cite{hinton2015distilling} is originally proposed to transfer knowledge from one model (\ie, teacher model) to another (\ie, student model) to compress the size of model. Inspired by recent approaches to transfer knowledge between equivalent models~\cite{furlanello2018born}, we propose using the inferior (\eg, less accurate) predecessor model as a teacher model to mitigate example forgetting of the student model (\ie, last trained, superior model) by learning from pseudo labels~(\ie{logits}). This distillation method can be interpreted as a type of consistency regularizer to alleviate forgotten knowledge.

Formally, Algorithm~\ref{alg:TrustAL} describes the overall procedure of the TrustAL framework on the AL iterations. When given a labeled data pool $\calL = \{x_i, y_i\}_{\forall i}$ at the $t$-th iteration, let $L_{CE}$ be a typical cross-entropy loss with oracle labeled examples, \ie, $\sum_{(x_i,y_i) \in \mathcal{L}} CrossEntropy(y_i, f(x_i;\theta_t))$,  and let $L_{KL}$ be a knowledge distillation loss with the pseudo labels of a predecessor model from $t\text{-}\Delta t$, \ie, $\sum_{(x_i,y_i) \in \mathcal{L}} KL\text{-}Divergence(f(x_i;\theta_{t-\Delta t}), f(x_i;\theta_t))$. On the top of an arbitrary data acquisition method (\eg, CORESET and BADGE), model parameter $\theta_t$ produced by TrustAL is obtained  by the following optimization:
\begin{equation} \label{eq:trustal}
    \theta_t = argmin_{\theta_t} L_{CE}(\theta_t) + \alpha \cdot L_{KL}( \theta_{t-\Delta t}, \theta_t)
\end{equation}
where $\alpha$ is a preference weight. We empirically analyze the effect of varying $\alpha$ in Appendix D.

This framework motivates us to leverage more sophisticated techniques for knowledge distillation, such as~\cite{yuan2020revisiting,park2019relational}. We leave such exploration for future work, as using Eq. (\ref{eq:trustal}) works quite well for multiple AL methods in our experiments. 
Instead, as any predecessor model can be a teacher, we extend this framework to further exploration of teacher selection.

\begin{algorithm}[t!]
    \fontsize{9.5}{11.5}\selectfont
    \caption{TrustAL}\label{alg:TrustAL}
    \STATE{\textbf{Input:} Initial labeled data pool $\calL_0$, unlabeled data pool $\calU$, number of queries per iteration (budget) $k$, number of iterations $T$, sampling algorithm $\calA$, fixed development dataset $\calD_{dev}$ with size $m$}\\
    \STATE{\textbf{Output:} Model parameters $\theta_T$}\\
    \STATE{$\theta_0 \leftarrow$ Train a seed model on \calL}\\
    \For{iteration $t=1,...,T$}{
        \STATE{$M_t(x) = f(x;\theta_{t-1})$}\\
        \STATE{$\calQ_t$ $\leftarrow$ Apply $\calA(x, M_t, k)$ for $\forall x \in \calU$}\\
        \STATE{$\theta_{t-\Delta t} \leftarrow$ TeacherSelection$(\theta_0, ..., \theta_{t-1}, \calD_{dev})$}\\
        \STATE{$\bar{\calQ}_t \leftarrow $ Label queries $\calQ_t$ by oracles and $f(x;\theta_{t-\Delta t})$}\\
        \STATE{$\calL \leftarrow \calL \cup \bar{\calQ}_t$}\\
        \STATE{$\calU \leftarrow \calU\setminus \bar{\calQ}_t$}\\
        \STATE{$\theta_{t} \leftarrow$ $\operatorname*{argmin}_{\theta_t} L_{CE}(\theta_t) + \alpha \cdot L_{KL}(\theta_{t-\Delta t}, \theta_t)$}
    }

    \STATE{\textbf{return}~ $\theta_T$}
\end{algorithm}

\begin{figure}[t]
\centering
    \includegraphics[width=\linewidth]{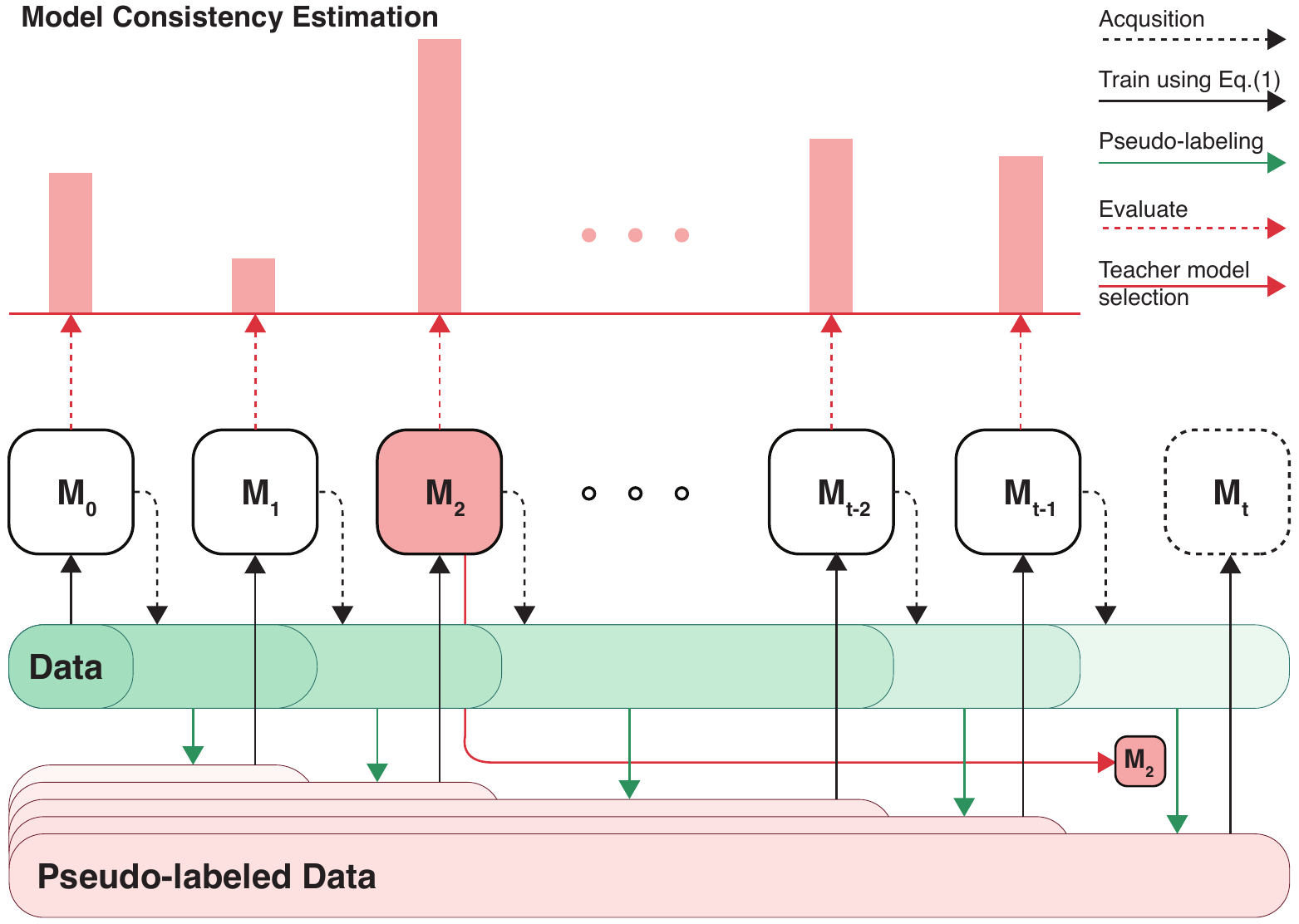}
\caption{Illustration of TrustAL-NC}
\label{fig:process-2}
\end{figure}

\begin{figure*}[t!]
     \centering
     \begin{subfigure}[b]{0.31\textwidth}
         \centering
         \includegraphics[width=\textwidth]{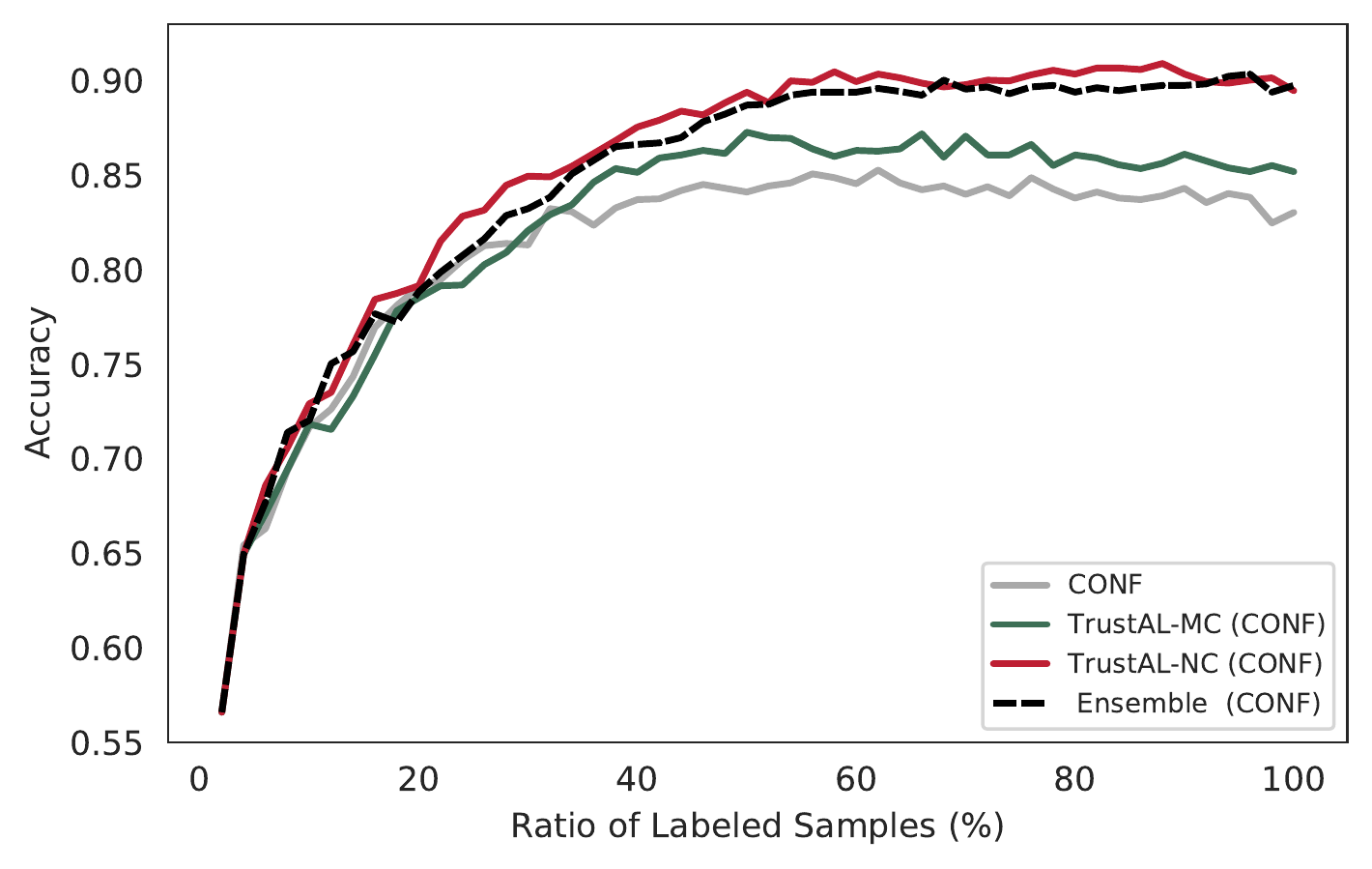}
         \caption{CONF}
     \end{subfigure}
     \hfill
     \begin{subfigure}[b]{0.31\textwidth}
         \centering
         \includegraphics[width=\textwidth]{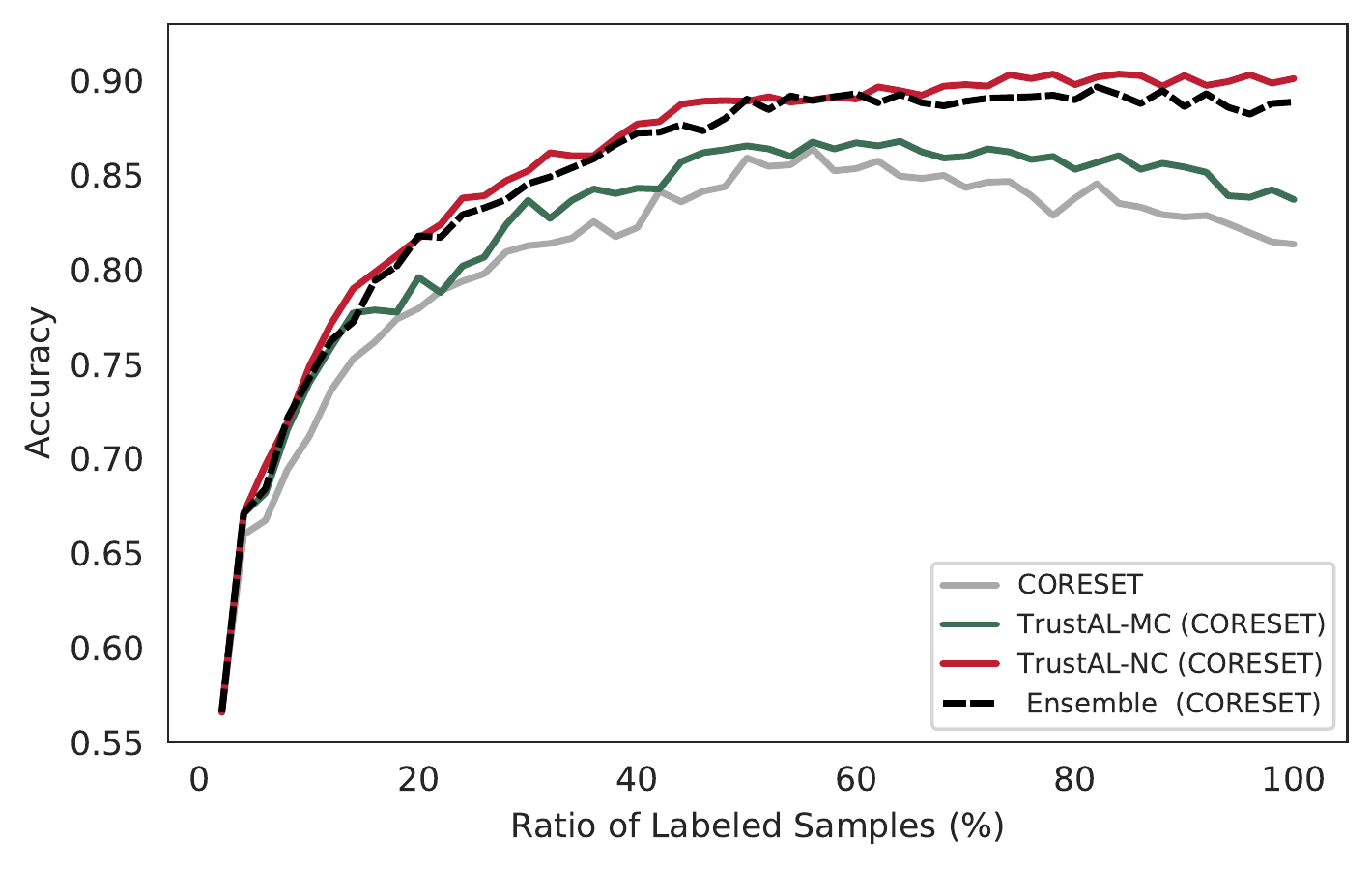}
         \caption{CORESET}
     \end{subfigure}
     \hfill
     \begin{subfigure}[b]{0.31\textwidth}
         \centering
         \includegraphics[width=\textwidth]{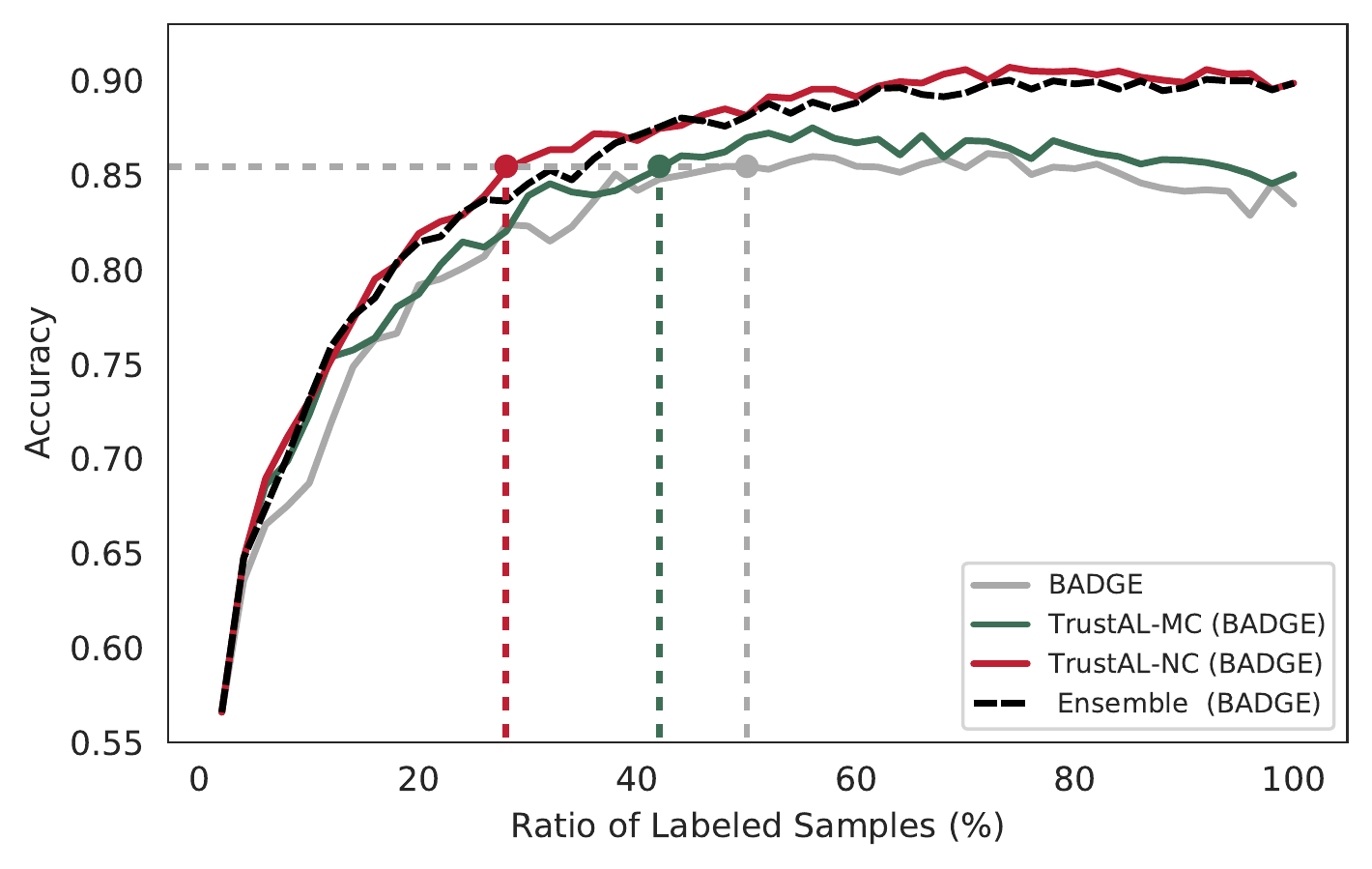}
         \caption{BADGE}
     \end{subfigure}
     \begin{subfigure}[b]{0.31\textwidth}
         \centering
         \includegraphics[width=\textwidth]{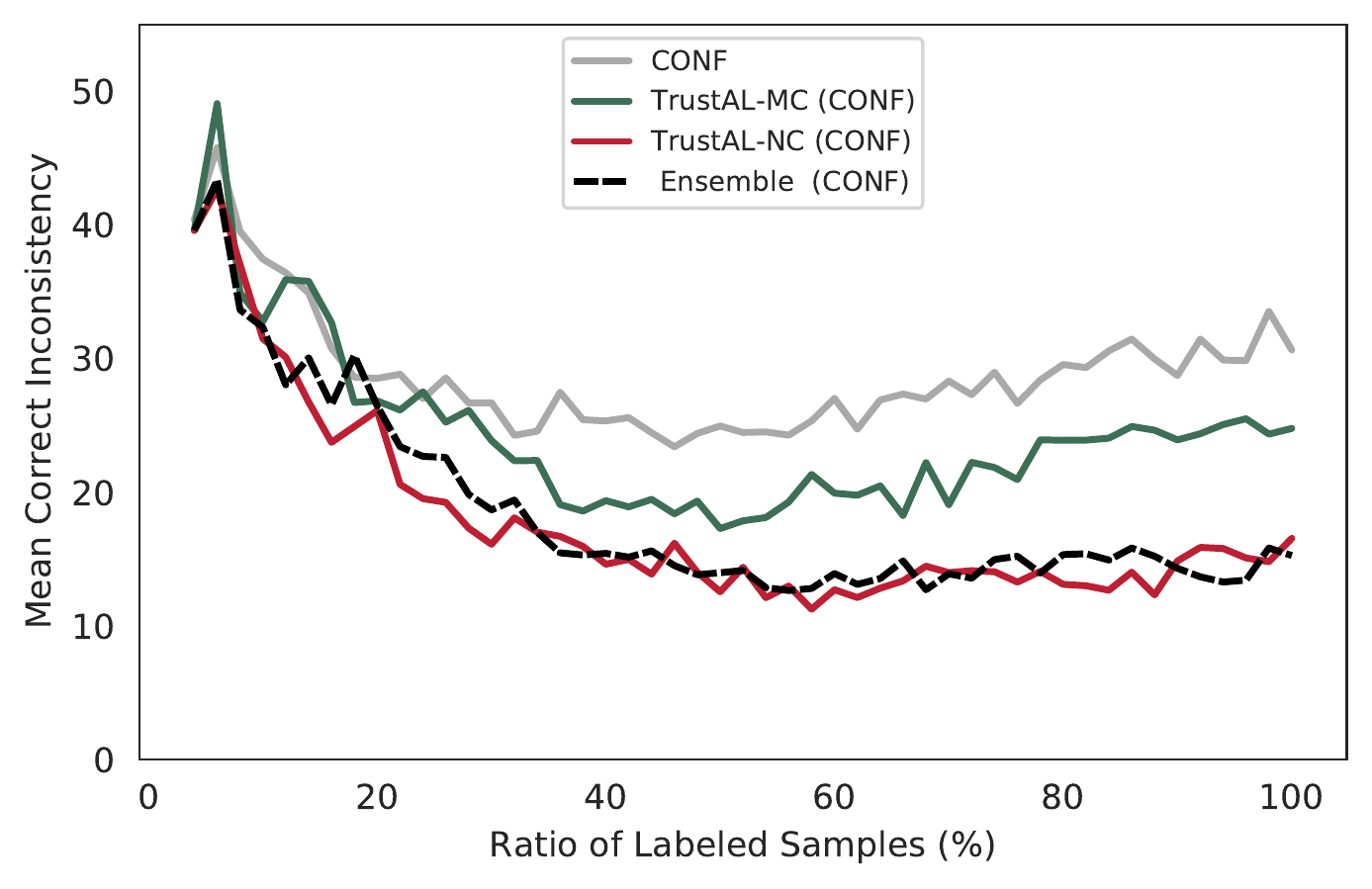}
         \caption{CONF}
     \end{subfigure}
     \hfill
     \begin{subfigure}[b]{0.31\textwidth}
         \centering
         \includegraphics[width=\textwidth]{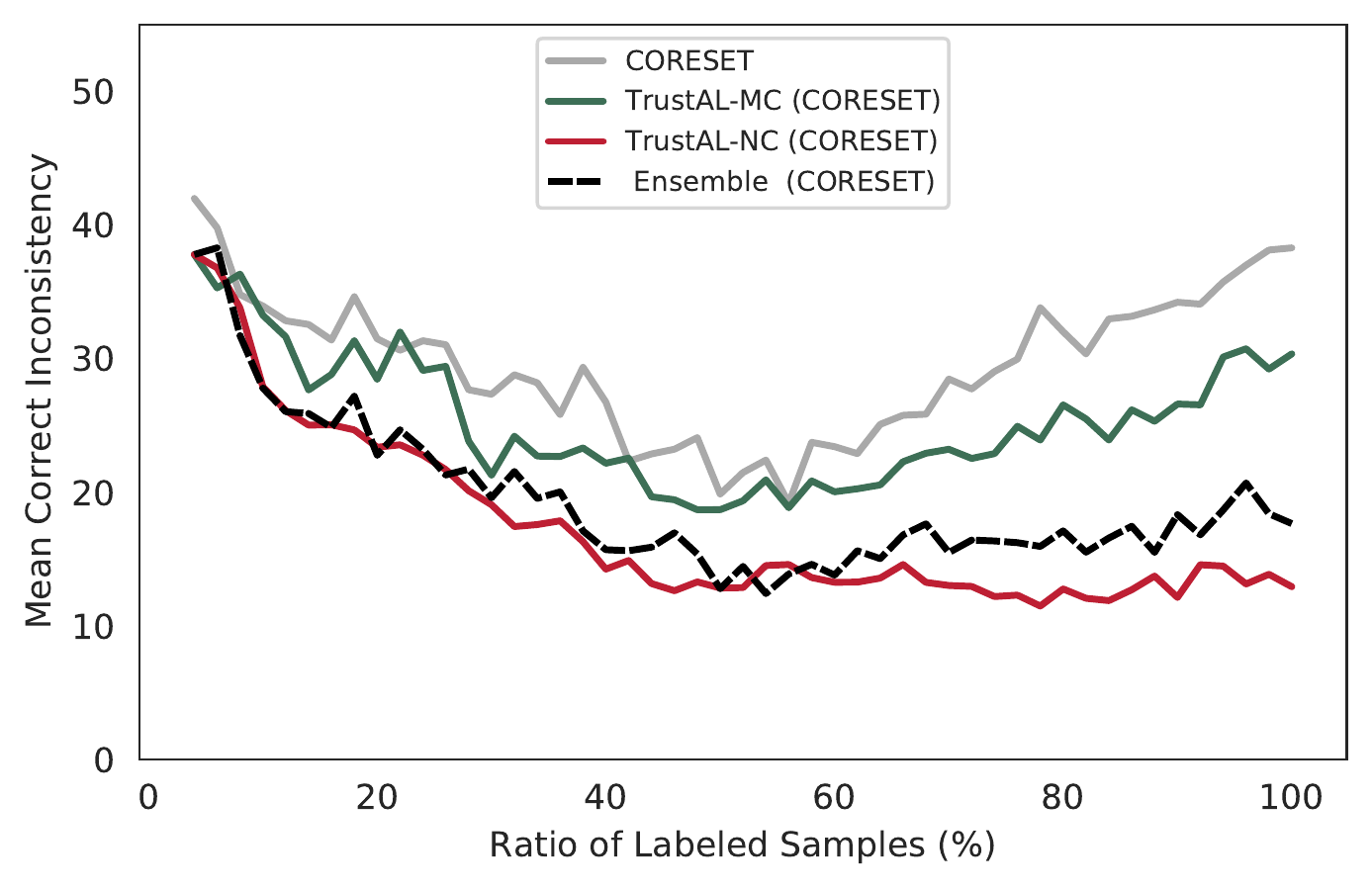}
         \caption{CORESET}
     \end{subfigure}
     \hfill
     \begin{subfigure}[b]{0.31\textwidth}
         \centering
         \includegraphics[width=\textwidth]{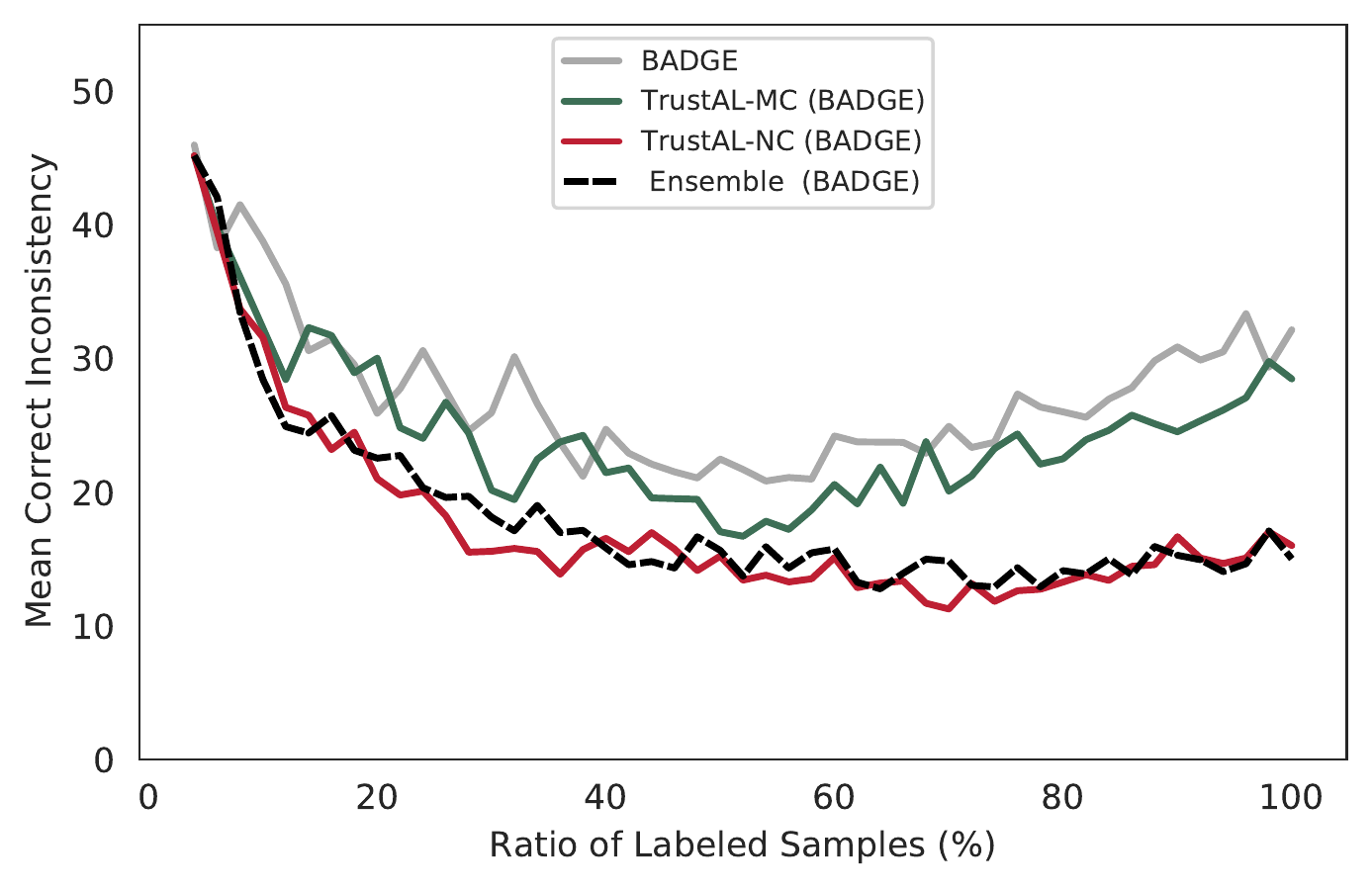}
         \caption{BADGE}
     \end{subfigure}
    \caption{Accuracy (a\text{-}c) and MCI (d\text{-}f) versus the ratio of labeled samples}
    \label{fig:rq1-Trec}
\end{figure*}

\subsection{Teacher Selection Strategies}\label{sec:trustal.2}
The key factor of TrustAL framework is considering a predecessor model as a specialist model for the forgettable knowledge. As reported in Figure~\ref{fig:exampleforgetting}, specific to data increments across multiple generations, predecessor models have different status of learned and forgotten knowledge. Therefore, the distillation effects are different in how to select teacher models. Here, we introduce two strategies with TrustAL: monotonic and non-monotonic consistency.

\subsubsection{Monotonic Consistency (TrustAL-MC)}
Basically, we can inherit the monotonic approach not only for data acquisition but also for teacher selection, synchronizing both, \ie, always $\theta_{t-\Delta t} = \theta_{t-1} = M_t$. This allows to iteratively transfer the learned knowledge generation by generation.

\subsubsection{Non-monotonic Consistency (TrustAL-NC)} \emph{Correct Inconsistency} (Definition~\ref{def:ci}) can be a strong signal to indicate which sample is forgettable for the current acquisition model. Using such sample level inconsistency, we aim at choosing the predecessor model with the learned knowledge especially of those forgettable samples. This allows to transfer the easily forgettable knowledge from one of the predecessor model, not always from $\theta_{t-\Delta t} = \theta_{t-1} = M_t$ as described in Figure~\ref{fig:process-2}.

Specifically, given a development dataset $\calD_{dev}$ with $m$ samples, let $\mathbb{C}^{t}$ be a vector of correct inconsistency values of $M_t$ ($=\theta_{t-1}$) at the $t$-th iteration for all $m$ samples, \ie, $ \langle\mathbb{C}_1^{(t-1)},...,\mathbb{C}_m^{(t-1)}\rangle \in \mathbb{R}^{m}$. For the purpose of using this vector as importance weights for samples, we normalize $\mathbb{C}^{t}$ into $\Tilde{\mathbb{C}}^{t}$ where $\sum_{\forall i} \Tilde{\mathbb{C}}_i^{t} = 1$ by a softmax function. We note that the sample $x_i$ with high importance weight $\Tilde{\mathbb{C}}_i^{t}$ means easily forgettable sample for $M_t$. Based on such consistency-aware sample importance, we define a function $g(\theta_{t-\Delta t},M_t)$ of measuring how reliably a predecessor model $\theta_{t-\Delta t}$ can be a synergetic teacher with the data acquisition of $M_t$, by an weighted accuracy as:
\begin{equation}\label{eq:estimation}
    g(\theta_{t-\Delta t},M_t) = {\Tilde{\mathbb{C}}^{t}}{}^{\top} \langle acc_1^{t-\Delta t},...,acc_m^{t-\Delta t}\rangle / m
\end{equation}
High $g(\theta_{t-\Delta t},M_t)$ implies that the teacher model $\theta_{t-\Delta t}$ tends to have the knowledge of forgettable examples for the current data acquisition model $M_t$, and vice versa. Based on this, we can select a predecessor having the maximum value, as a teacher model to teach a new model $\theta_t$:
\begin{equation}\label{eq:estimation}
    \operatorname*{argmax}_{1 < \Delta t \leq t } g(\theta_{t-\Delta t},M_t)
\end{equation}

\subsection{Development Set Strategies}
One of the plausible tools to estimate the learning status of AL generations is development set as it is often used for validation process. In fact, TrustAL-NC catches forgetting signals as a by-product of the validation process. The experiment on the robustness of TrustAL-NC on the size of development set shows marginal performance decrease even when halving development set size, which resolve concerns about keeping development set in label-scarce situation. Full empirical results are presented in Appendix D.

\section{Experiments}
\label{sec:experiment}


\subsection{Experimental Setup}
\label{sec:experiment.1}

\subsubsection{Dataset} We use three text classification datasets, TREC~\cite{roth2002question}, Movie review~\cite{pang2005seeing} and SST-2~\cite{socher2013recursive}, which are widely used in AL~\cite{lowell2018practical,siddhant2018deep,yuan2020cold} and statistically diverse. 
The data statistics are presented in Appendix A for more details.

\subsubsection{Baselines} As TrustAL is orthogonally applicable to any data acquisition strategy, for the purpose of better analysis, we use the following three acquisition methods as baselines. 
\begin{itemize}
    \setlength\itemsep{0.0005em}
    \item \textbf{CONF}~\cite{wang2014new}: An uncertainty-based method that selects samples with least confidence.
    
    \item \textbf{CORESET}~\cite{sener2017active}: A diversity-based method that selects coreset of remaining samples. 
    
    \item \textbf{BADGE}~\cite{ash2019deep}: A hybrid method that selects samples considering both uncertainty and diversity.
\end{itemize}
More details on the baselines are discussed in Appendix B.

\subsubsection{Implementation}
For all three datasets, we follow the commonly used default settings in AL for text classification~\cite{liu2021deep,zhou2021towards,lowell2018practical,siddhant2018deep}: Bi-LSTM~\cite{hochreiter1997lstm} is adopted as a base model architecture; In each iteration of AL, training a classification model from scratch (not by incremental manner) with the entire labeled samples gathered, to avoid the training issues with warm-starting~\cite{ash2020warm}. Note that the development set is held out in every experiments so that it is not used for training models. We describe our implementation details in Appendix C.

\begin{figure}[t!]
     \centering
     \begin{subfigure}[t]{0.23\textwidth}
         \includegraphics[width=\textwidth]{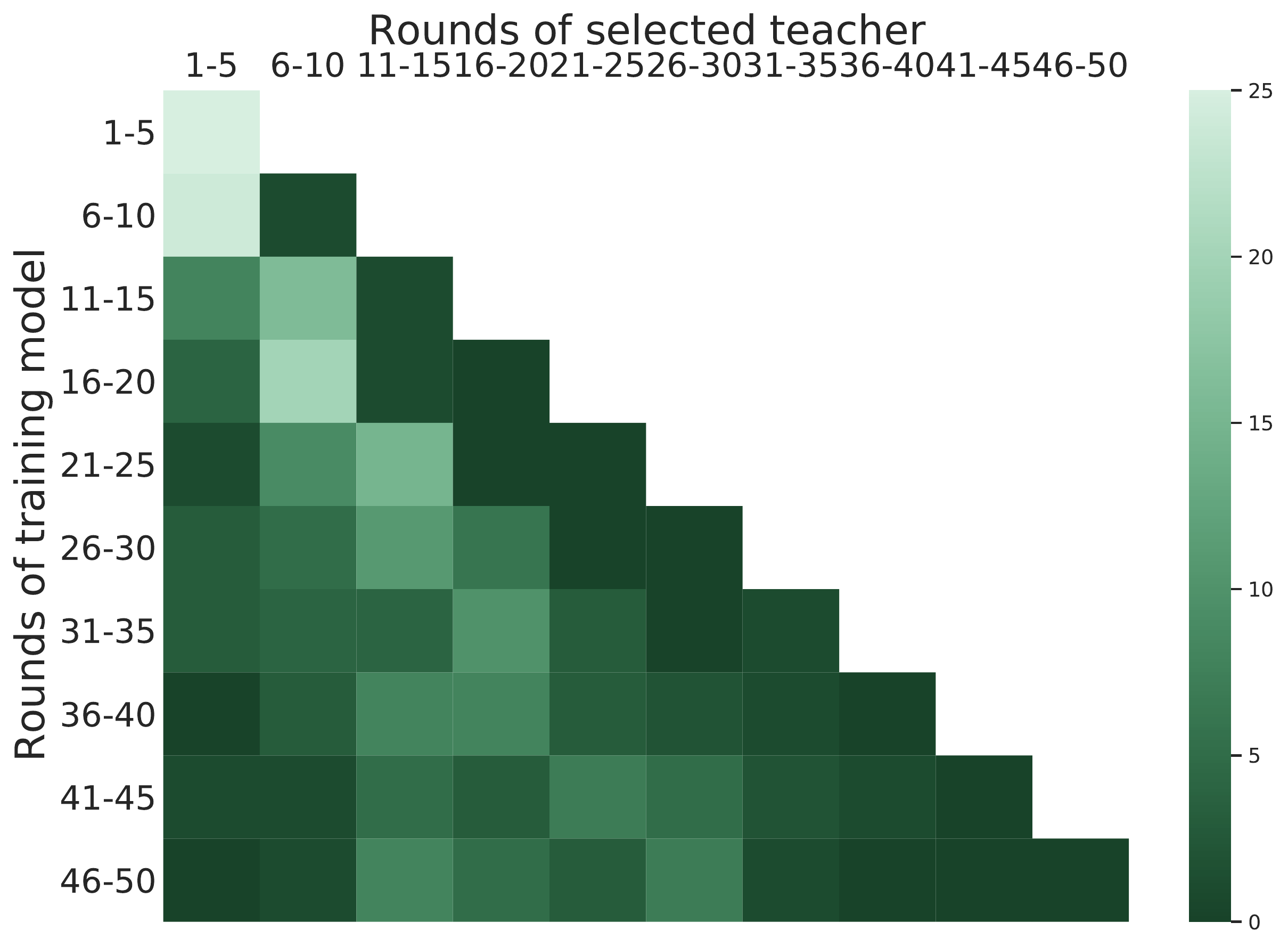}
         \caption{TREC}
         \label{fig:sss}
     \end{subfigure}
     \hfill
     \begin{subfigure}[t]{0.23\textwidth}
         
         \includegraphics[width=\textwidth]{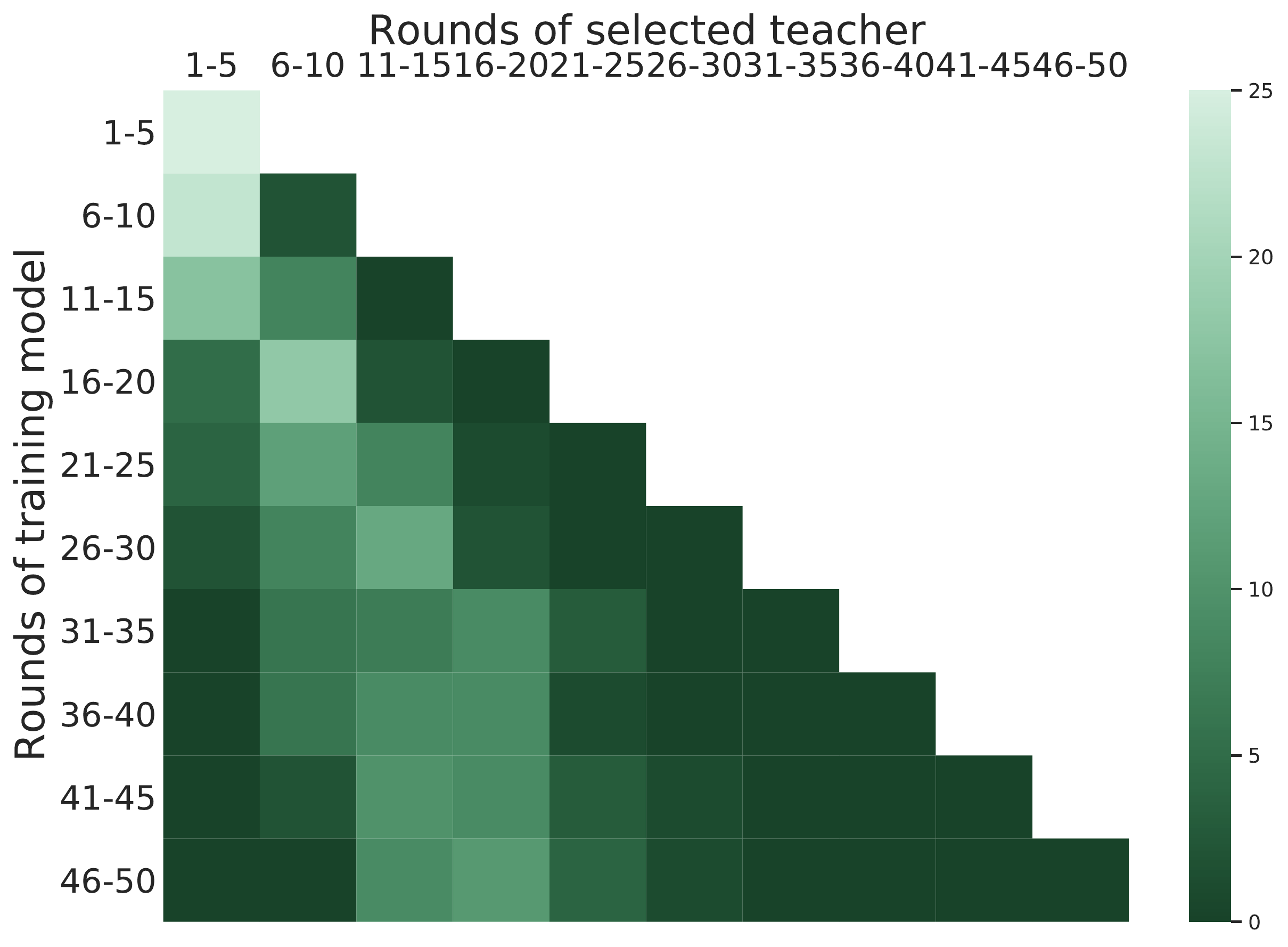}
         \caption{Movie Review}
     \end{subfigure}

     \begin{subfigure}[t]{0.23\textwidth}
         
         \includegraphics[width=\textwidth]{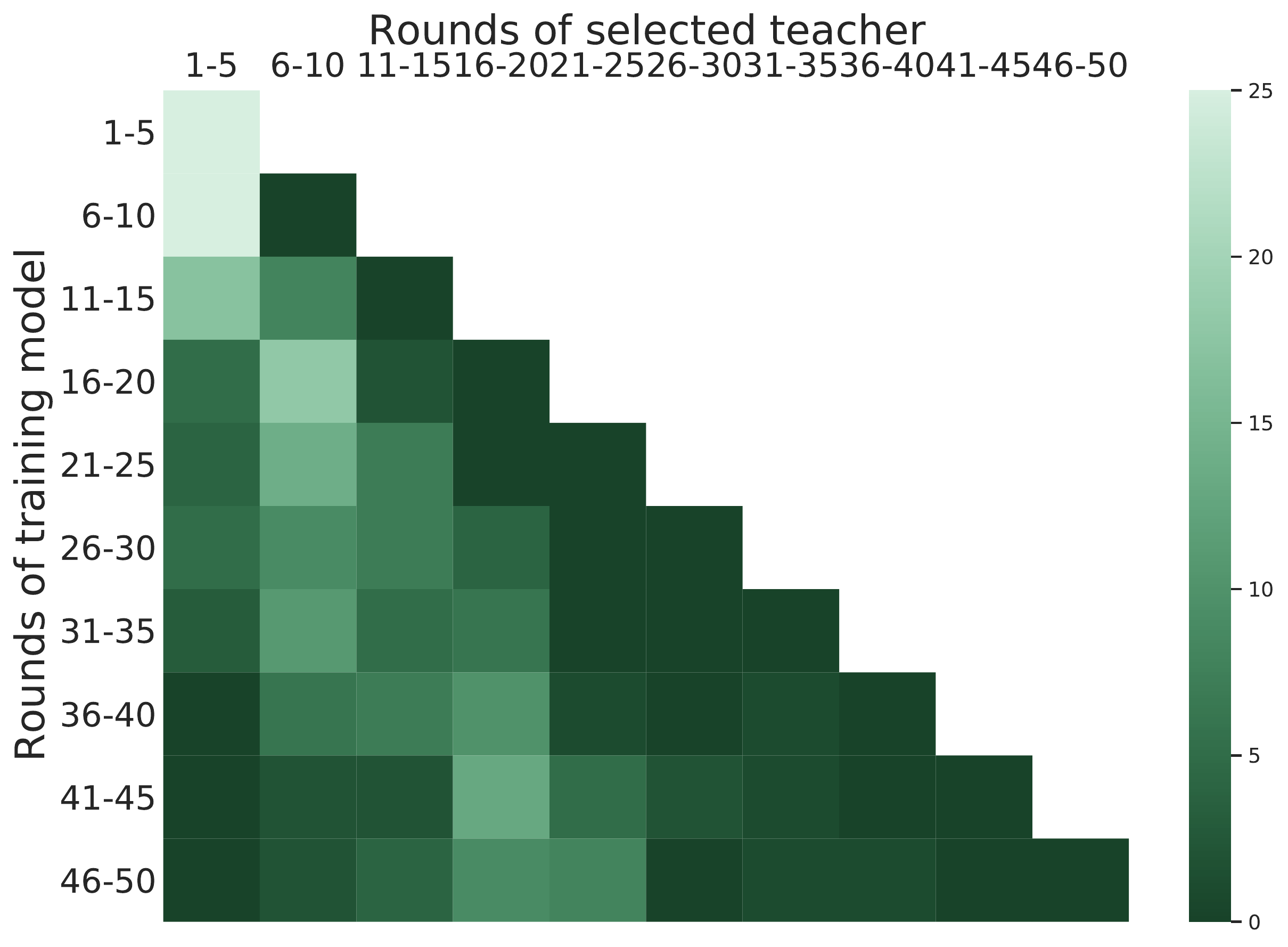}
         \caption{SST-2}
     \end{subfigure}
     \hfill
     \begin{subfigure}[t]{0.23\textwidth}
         
         \includegraphics[width=\textwidth]{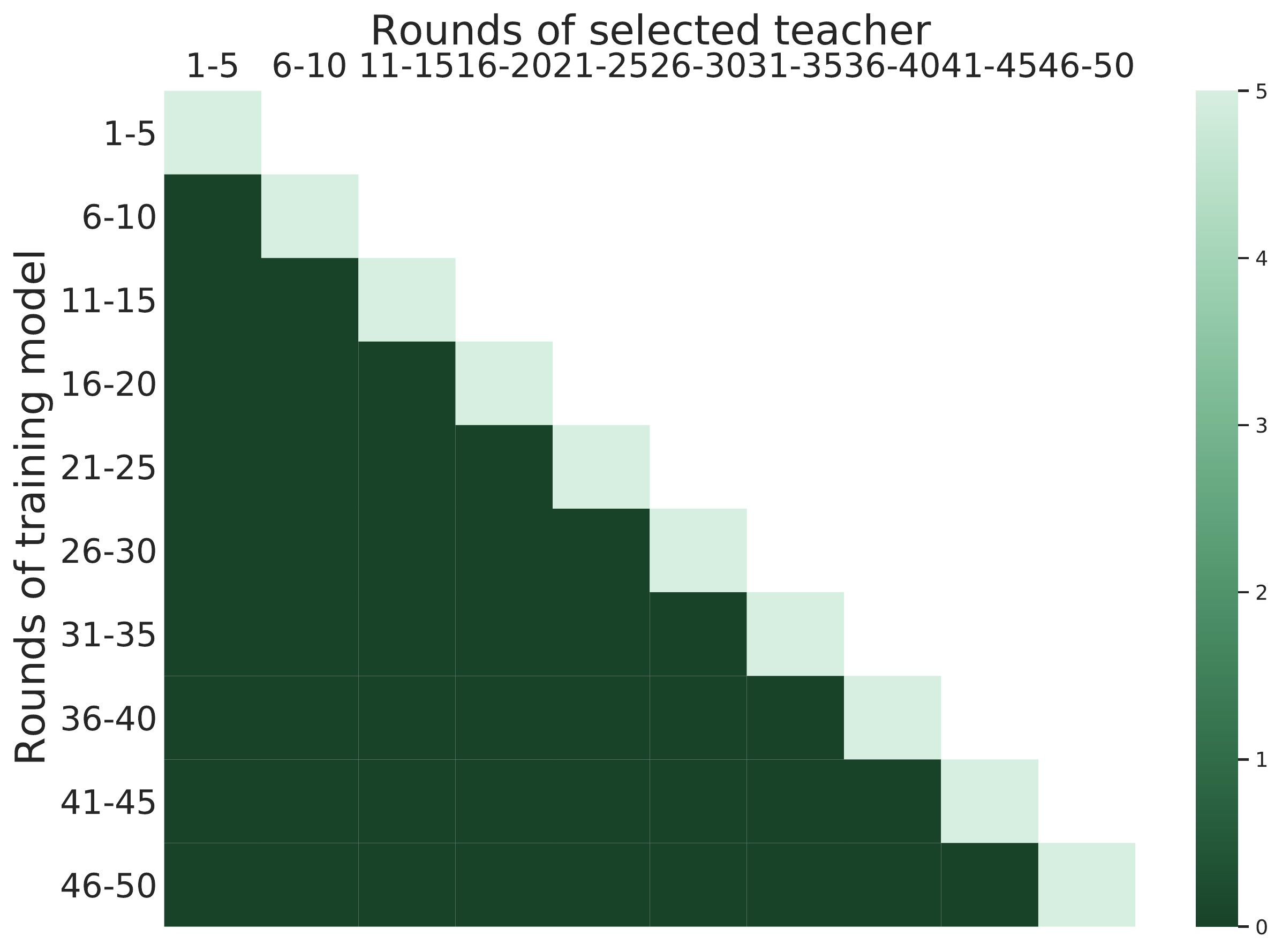}
         \caption{Trust-MC on all dataset}
         \label{fig:sss}
     \end{subfigure}
    \caption{Heatmap showing frequency of the model selection in TrustAL-NC with BADGE on three datasets. The $x$- and $y$-axis indicate the bins of the selected teacher model and the training model, respectively. In each round the size of labeled sample increases by 2\%. Each bin consists of 5 rounds. Brighter cell indicates more frequently selected bin.}
    \label{fig:heatmap}
\end{figure}


\subsection{Results and Discussion}\label{sec:experiment.2}

We present the empirical findings for the following three research questions:

\noindent\textbf{RQ1: } Does TrustAL outperform AL baselines? 

\noindent\textbf{RQ2: } How does TrustAL help data acquisition?

\noindent\textbf{RQ3: } Does TrustAL make consistent and robust models?\\
Additional experiments on hyperparameter sensitivity are presented in Appendix D.

\subsubsection{Overall Performance (RQ1)}\label{sec:new_rq_1}
First, we compare the performance of AL methods across AL iterations with and without TrustAL. Figure~\ref{fig:rq1-Trec} shows accuracy and MCI of AL methods on TREC. The empirical results in SST-2 and Movie Review are presented in Appendix F.

Overall, AL strategies combined with TrustAL-NC/MC show improved label efficiency and relieved MCI compared to stand-alone baselines in all datasets. The models trained with TrustAL framework require much smaller number of labeled samples to achieve the same level of accuracy than baselines. To facilitate the comparison of label efficiency of TrustAL and a baseline, we draw the horizontal reference line where the baseline starts to show convergence in Figure~\ref{fig:rq1-Trec} (c). As a result, we find that TrustAL-MC and TrustAL-NC require only 40\% and 30\% of the training data pool, respectively, while baselines requires 50\% of total training data to reach the same level of accuracy. This result suggests that keeping consistency of model in AL is an essential criterion, and TrustAL successfully satisfies the ultimate goal of AL, \ie, improving the label efficiency.

Further, TrustAL-NC performs comparably to ensemble based distillation method \cite{Fukuda2017EfficientKD} which aims to distill the ensembled (i.e. averaged) probability distribution of multiple models. This indicates TrustAL-NC selects a teacher model that can effectively relieve forgotten knowledge, even without using all predecessor models. Figure~\ref{fig:heatmap} visualizes the behavior of teacher selection by TrustAL. The figure shows that TrustAL-MC selects the most recent model as its definition and TrustAL-NC chooses the teacher model based on the consistency guidance. While preferring the more generalized teacher models from the end of the stable learning stages (16-20) rather than earlier stages, TrustAL-NC also selects earlier generation that might be inferior but professional in terms of forgotten knowledge. That is, TrustAL-NC can select complementary models for forgotten knowledge in an automatic manner.


\begin{figure}[t!]\small
    \centering
     \begin{subfigure}[b]{0.3\textwidth}
         \includegraphics[width=\textwidth]{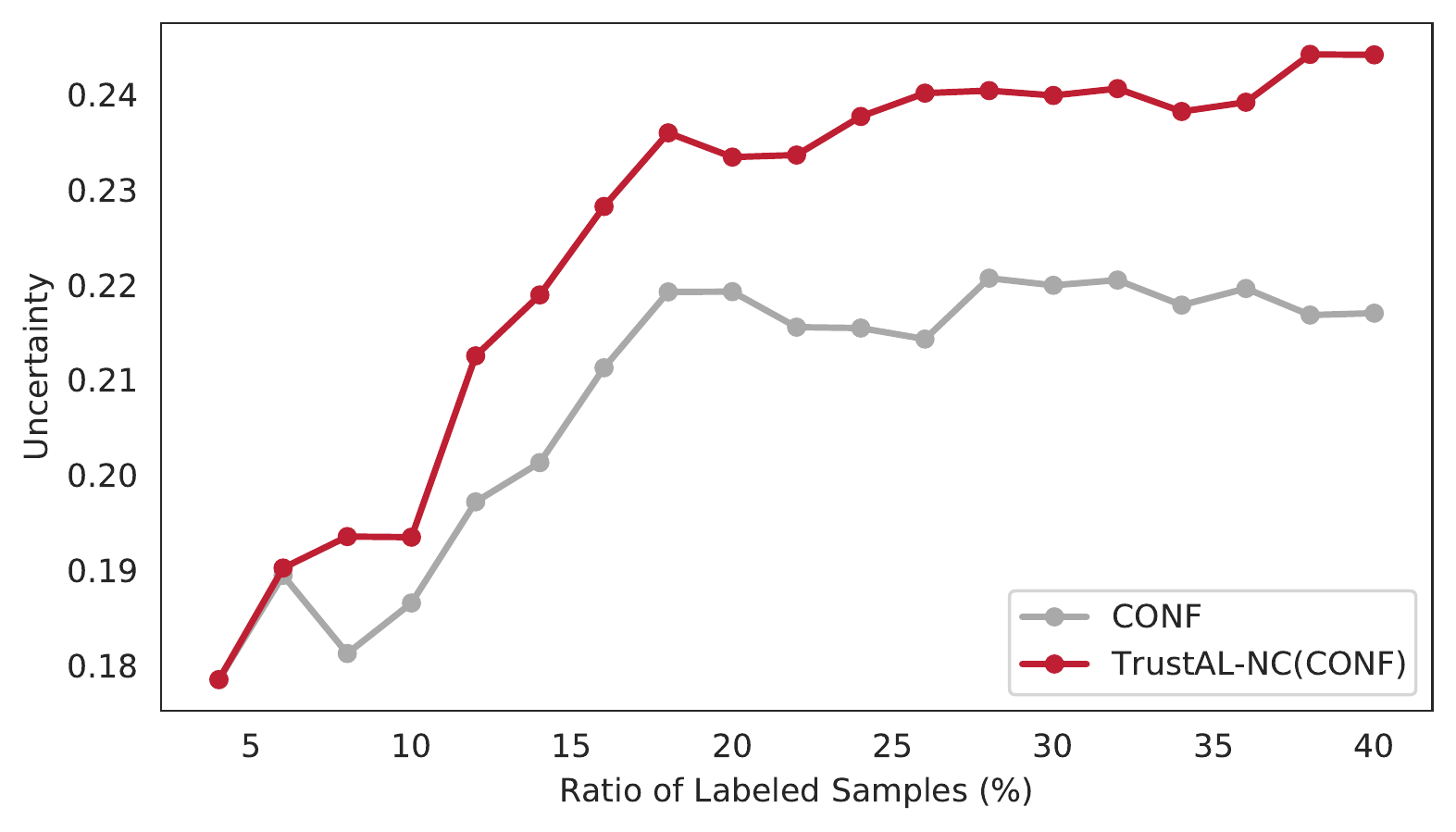}
         \caption{CONF}
         \label{fig:conf}
     \end{subfigure}
     \hfill\\
     \begin{subfigure}[b]{0.30\textwidth}
         \includegraphics[width=\textwidth]{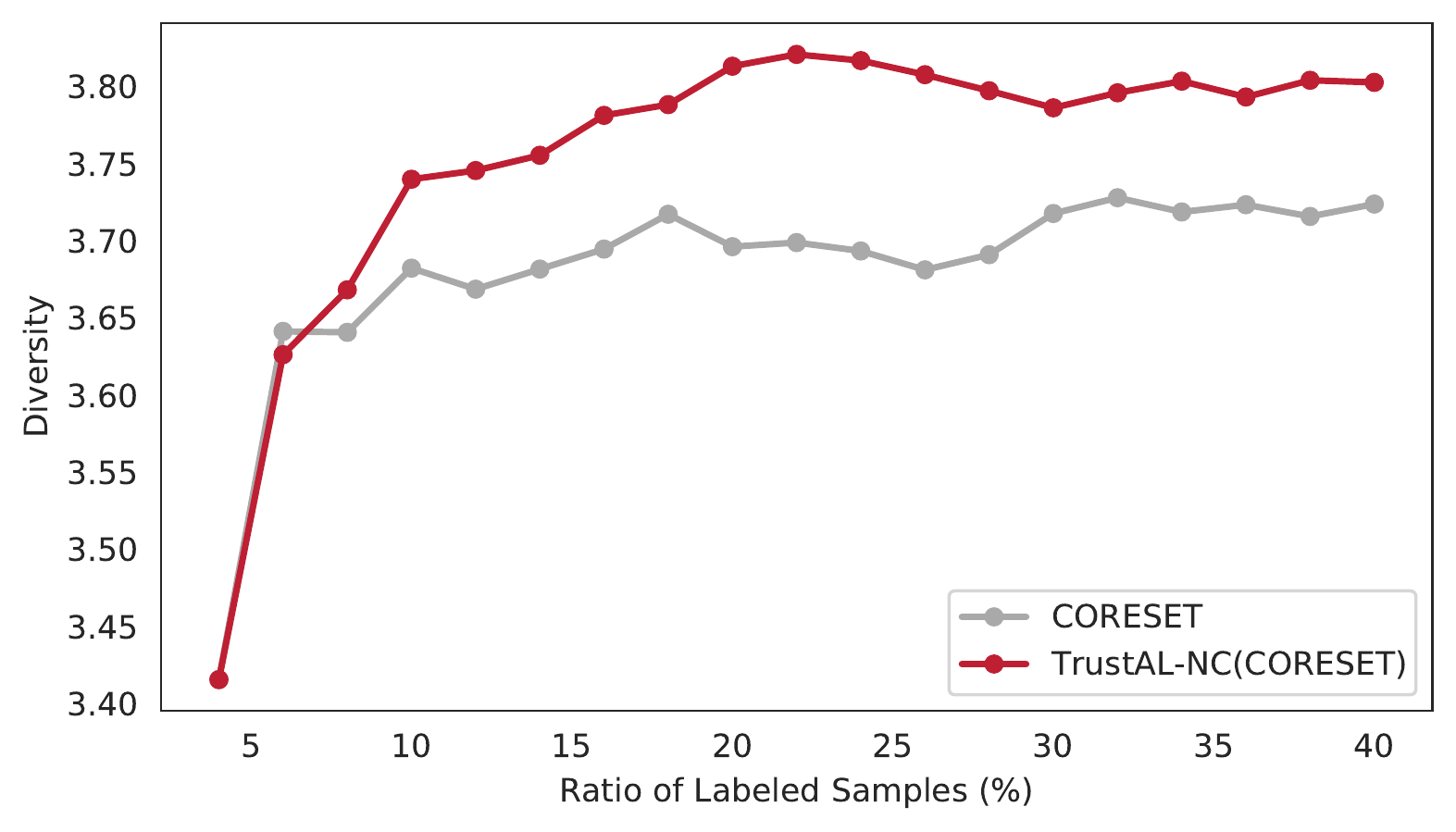}
         \caption{CORESET}
         \label{fig:div}
     \end{subfigure}
    \caption{Data acquisition analysis in stable phase on TREC; x-axis represents the ratio of labeled samples and y-axis represents the corresponding metrics.
    } \label{fig:rq2}
\end{figure}

\begin{table}[t!]\small
\setlength\extrarowheight{3pt}
\centering
 \begin{tabular}{c l c c c}
 \hline
 \noalign{\hrule height 0.8pt}
    \multicolumn{2}{c}{} & \textbf{TREC} & \textbf{Movie review}  & \textbf{SST-2} \\
\hline
\noalign{\hrule height0.8pt}
\multirow{4}{*}{A}    & baseline   & 0.726 & 0.637 & 0.686 \\\cline{2-5}
                         & Ensemble & 0.770 & 0.669 & 0.727 \\\cline{2-5}
                         & TrustAL-MC & 0.743 & 0.654 & 0.705 \\\cline{2-5}
                         & TrustAL-NC   & \textbf{0.774} & \textbf{0.676} & \textbf{0.730} \\\cline{1-5}
\multirow{4}{*}{B} & baseline & 0.727 & 0.627 & 0.697 \\\cline{2-5}
                         & Ensemble   & 0.777 & 0.658 & 0.724 \\\cline{2-5}
                         & TrustAL-MC & 0.753 & 0.654 & 0.707 \\\cline{2-5}
                         & TrustAL-NC   & \textbf{0.785} & \textbf{0.665} & \textbf{0.724} \\\cline{1-5}
\multirow{4}{*}{C}   & baseline   & 0.735 & 0.636 & 0.681 \\\cline{2-5}
                         & Ensemble   & 0.773 & 0.668 & 0.722 \\\cline{2-5}
                         & TrustAL-MC & 0.748 & 0.653 & 0.711 \\\cline{2-5}
                         & TrustAL-NC & \textbf{0.780} & \textbf{0.670} & \textbf{0.729} \\\cline{1-5}
 \hline
 \noalign{\hrule height0.8pt}
 \end{tabular}
\caption{Correct consistency of TrustAL-NC with (A) CONF (B) CORESET and (C) BADGE} 
\label{table:ap.wangcorrectconsistency}
\end{table}

\subsubsection{Data Acquisition Quality (RQ2)}\label{sec:new_rq_2} Having tested for the overall accuracy and MCI of TrustAL, we evaluate the quality of data acquisition results when using TrustAL. To discuss how TrustAL affects data acquisition, we analyze TrustAL-NC based on the two distinctive strategies on data acquisition: uncertainty and diversity. Note that, we choose to compare acquisition quality of stable phase only since the label efficiency of saturated phase is negative for traditional data acquisition strategies in AL.

For uncertainty, following~\cite{yuan2020cold}, we first obtain a reference model trained on the full training data of a target task, then measure the uncertainty of samples selected on each iteration. Specifically, by using Shannon Entropy~\cite{shannon2001mathematical}, we compute the entropy of the predicted probability distribution of individual samples, and report their average values for each AL iteration in Figure~\ref{fig:conf}, where a higher value implies each iteration successfully acquires uncertain samples.

For diversity, we reuse the reference model to encode the full training data into a feature space, then obtain the $k$ disjoint sets of the all training data by K-means algorithm where we set $k$ as the number of samples acquired per AL iterations. Then, based on these $k$ groups, we measure the diversity of a sample set selected on each iteration, by computing the entropy of a cluster distribution of the selected samples, which we report in Figure~\ref{fig:div}. The measure shows whether the samples are uniformly picked among the clusters, since diversely acquired samples would belong to different clusters~\cite{ash2019deep}.


As shown in Figure~\ref{fig:rq2}, providing more labeled data leads to improvement of uncertainty and diversity for both baseline and TrustAL. Considering the reported increase of generalization performance in RQ1, this suggests that better model training leads to better acquisition, strengthening model's ability to identify more informative samples. For CONF and CORESET each representing uncertainty and diversity based strategies, we observe that TrustAL largely improves the quality of acquisition across the AL procedure. Since TrustAL aims to enhance the model's ability of surrogating the labeled dataset, we believe that TrustAL can be orthogonally and effectively applicable to any acquisition strategy with its synergetic nature.

\subsubsection{Model Consistency and Robustness (RQ3)}\label{sec:new_rq_3}

The correct consistency~\cite{wang2020wisdom}, $\ie, avg_{\forall i} \mathbbm{1}_{\hat{y}^m_i = \hat{y}^n_i = y_i}$, is a measure for the consistency between $m^{th}$ and $n^{th}$ generation models. By measuring the correct consistency between any two models, we demonstrate that our framework contributes not only to the label efficiency, but also to the overall consistency of model generations. Shown in Table~\ref{table:ap.wangcorrectconsistency}, TrustAL-NC shows better correct consistency than baselines. This reveals that models in AL iterations accord with each other for the correctly classified samples, which is also related to user's trust on system~\cite{wang2020wisdom}.

\begin{figure}[t]
     \centering
     \begin{subfigure}[t]{0.23\textwidth}
         \includegraphics[width=\textwidth]{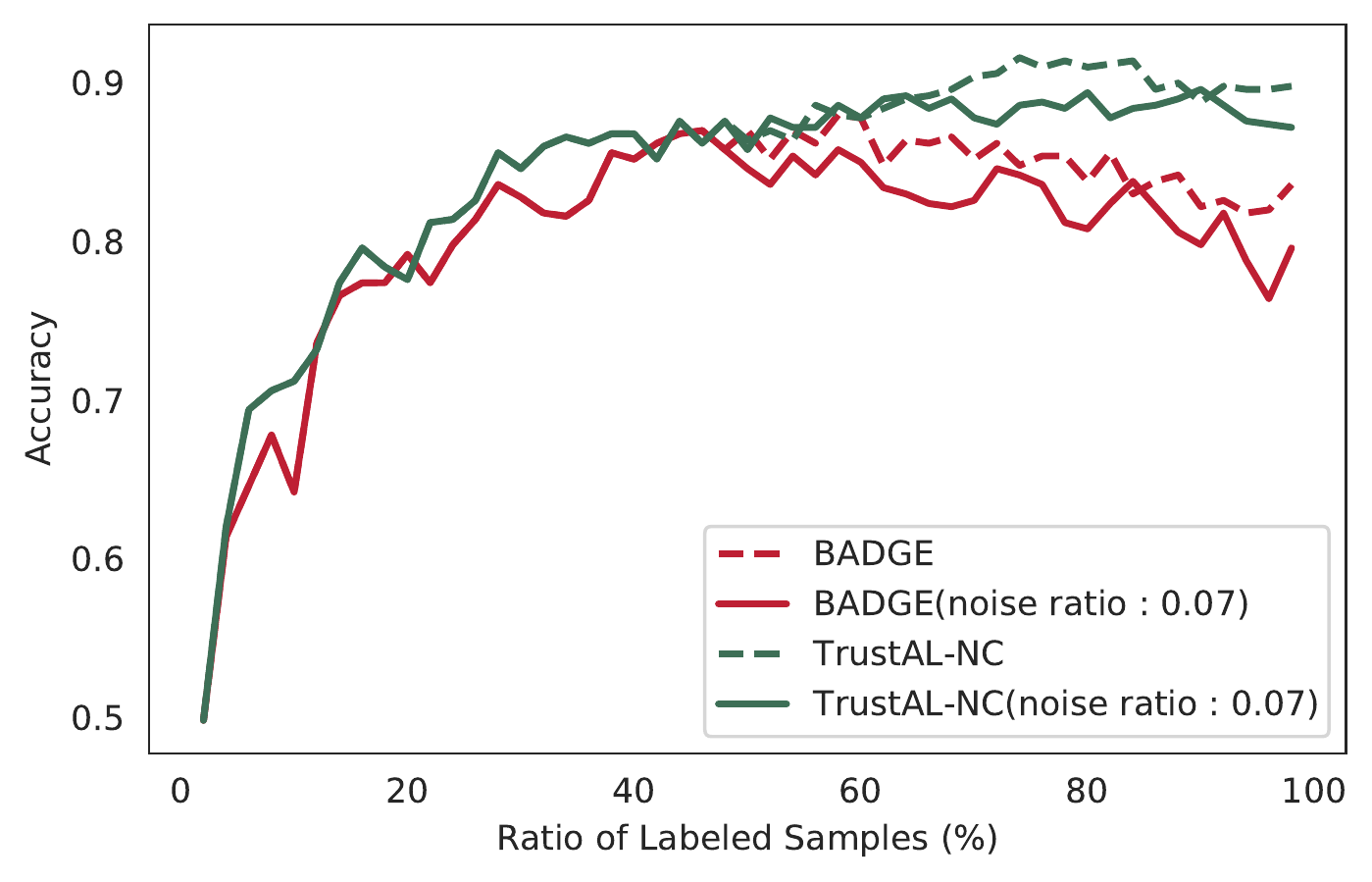}
         \label{fig:sss}
     \end{subfigure}
     \hfill
     \begin{subfigure}[t]{0.23\textwidth}
         
         \includegraphics[width=\textwidth]{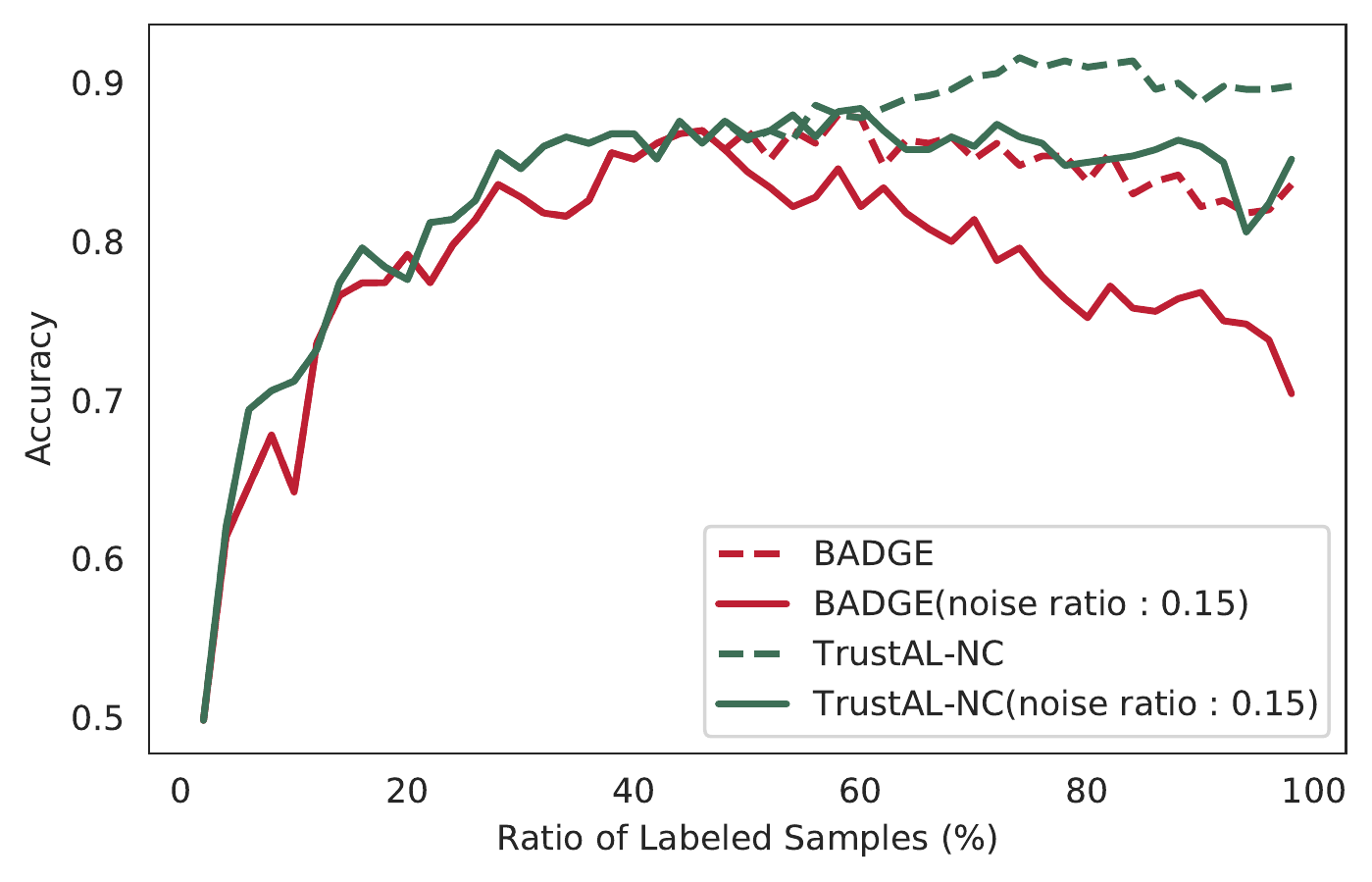}
     \end{subfigure}

     \begin{subfigure}[t]{0.23\textwidth}
         
         \includegraphics[width=\textwidth]{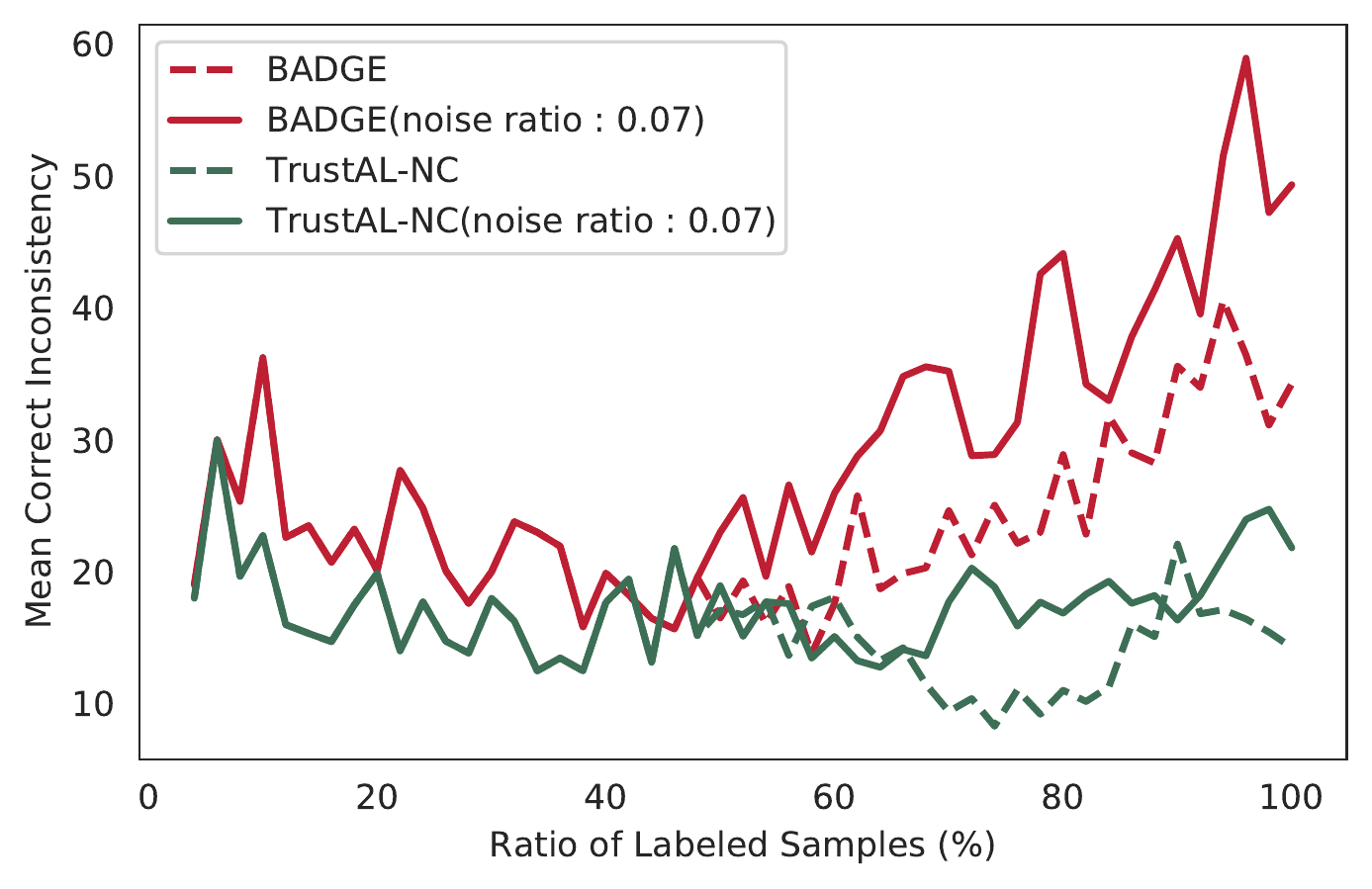}
         \caption{Noise ratio : 7\%}
     \end{subfigure}
     \hfill
     \begin{subfigure}[t]{0.23\textwidth}
         \includegraphics[width=\textwidth]{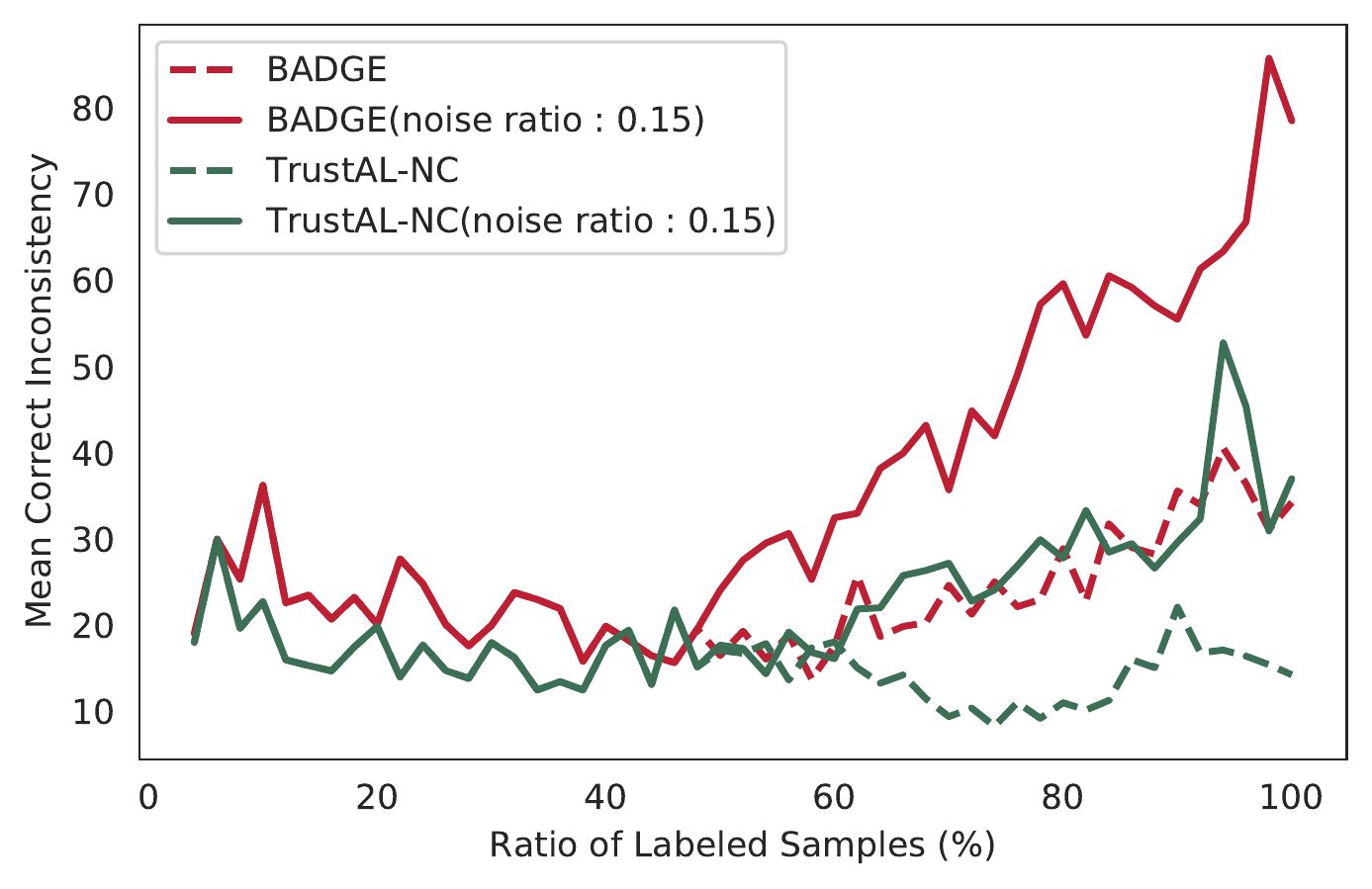}
         \caption{Noise ratio : 15\%}
         \label{fig:sss}
     \end{subfigure}
    \caption{Robustness analysis varying the ratio of noise. Accuracy and MCI are shown in pair for each noise ratio.}
    \label{fig:noisy_label}
\end{figure}

To demonstrate the robustness of TrustAL on the careless transition from human labeling, we deliberately corrupt the acquired samples by randomly flipping certain ratio of labels. Specifically, after stable phase, we corrupt 7\% and 15\% of the labels. In Figure~\ref{fig:noisy_label}, the result of BADGE and TrustAL(BADGE) are shown. Since other strategies show similar behaviors, we only present the result of BADGE. When the stable phase ends, the noisy labels cause the rapid increase of forgotten knowledge. Based on this observation, we believe that one of the possible suspects of performance degradation in the saturated phase might be noisy examples. Despite such degradation, TrustAL performs more robustly which is in strike contrast to the baseline. With 7\% of noise, TrustAL even shows comparable result to the one trained without noise. This result shows that TrustAL is robust to accidental noise in labels produced by human annotators since TrustAL regularizes the negative impact of such labels by pursuing consistency as an additional objective in training.



\section{Conclusion} \label{sec:conclusion}



In this paper, we debunk the monotonicity assumption which is a common belief in conventional AL methods by empirical observation of example forgetting. For that, we present TrustAL, an effective and robust framework that uses the predecessor model as an expert model for knowledge distillation to compensate the loss of knowledge between data and model. Especially, our framework can be orthogonally applicable to existing data acquisition in a highly efficient way. Further, we present multi-pronged analysis for our method through extensive experiments. 

\section{Acknowledgements}
This work was partly supported by Institute of Information \& communications Technology Planning \& Evaluation (IITP) grant funded by the Korea government(MSIT) (No. 2020-0-01361, Artificial Intelligence Graduate School Program (Yonsei University)). And this research was partly supported by the MSIT(Ministry of Science, ICT), Korea, under the High-Potential Individuals Global Training Program)(2021-11-1603) supervised by the IITP(Institute for Information \& Communications Technology Planning \& Evaluation). Jinyoung Yeo is a corresponding author.
\bibliography{aaai22.bib}
\clearpage

\appendix

\section{Appendix A: Dataset} \label{sec:ap.dataset} 

We use the following three text classification datasets. The statistics for each dataset is shown in Table ~\ref{table:ap.Stat-Text-Classification}.
We split the official training split into 90\% and 10\% as training and validate sets respectively. 

\vspace{3pt}
\noindent\textbf{TREC~\cite{roth2002question}:} This dataset is for question classification task with 6 categories related to the subject of the question. 

\vspace{3pt}
\noindent\textbf{Movie Review~\cite{pang2005seeing}:} This polarity dataset is for the sentimental classification task (positive/negative) using movie review snippets on a five start scale.

\vspace{3pt}
\noindent\textbf{SST-2~\cite{socher2013recursive}:} Stanford Sentiment Treebank dataset is for sentimental classification task using a binary class (positive/negative). 

\begin{table}[h]\small
\setlength\extrarowheight{3pt}
\centering
 \begin{tabular}{l c c c c}
 \hline
 \noalign{\hrule height0.8pt}
\textbf{Dataset} & \textbf{Train}  & \textbf{Dev} & \textbf{Test} & \textbf{\#Classes}\\
 \hline
\textbf{TREC}  &4,952  & 500 & 500 &6\\
\textbf{Movie review}  & 8,526 & 1,068 & 1,068& 2\\
\textbf{SST-2} & 6,920 & 872 &  1,821 & 2 \\
 [0.5ex]
 \hline
 \noalign{\hrule height0.8pt}
 \end{tabular}
\caption{Text classification dataset statistics}
\label{table:ap.Stat-Text-Classification}
\end{table}

\section{Appendix B: Baselines} \label{sec:ap.baselines} 
We apply the following three data acquisition methods to TrustAL framework, while considering their stand-alone as baselines.

\vspace{3pt}
\noindent\textbf{CONF}~\cite{wang2014new}: This strategy samples with the smallest of the predicted class logit, for k classes, $max_{i=1}^k f(x;\theta)_i$. Specifically, the sample is classified as uncertain if the probability of the most probable label for a sample is small.

\vspace{3pt}
\noindent\textbf{CORESET}~\cite{sener2017active}: This approach attempts to select samples by constructing a core subset. The embedding of each sample is computed by the second end layer and the samples at each round are selected using a greedy furthest-first traversal conditioned on all labeled samples.

\vspace{3pt}
\noindent\textbf{BADGE}~\cite{ash2019deep}: The approach samples point by using clustering in the hallucinated gradient space represented approximately based on pseudo labels. This method automatically balances the prediction uncertainty of the model and the diversity of the samples in a batch.

\section{Appendix C: Training Details} \label{sec:ap.hyperparameter} 
In the entire process of training model and conducting experiments, we used machine with Intel CoffeeLake i5-9400F CPU, 64GB of RAM and RTX 2080ti GPU. For the model architecture, we use single layer Bi-LSTM with 200 hidden units with dropout 0.5. The hyperparameters for LSTMs on each task are listed in Table~\ref{table:ap.lstm-hyper}.

We conduct the experiments in section~\ref{sec:new_rq_1} with five random seeds. We represent words initially with GloVe vectors~\cite{pennington-etal-2014-glove} with 300-dimensional glove embeddings pretrained on 6B tokens. We use a batch size set to be 50 which is common setting in AL assuming low budget. The active learning process begins with a random acquisition of 2\% from the dataset. We train an initial model on this data. Then, we iteratively apply baseline acquisition function to sample an additional 2\% of examples and train a new model based on this data. For each round, we evaluate models for every epoch on development set and select the model with the highest validation accuracy to report the test accuracy.

\begin{table}[h]\small
\setlength\extrarowheight{3pt}
\centering
 \begin{tabular}{c c c c}
 \hline
 \noalign{\hrule height 0.8pt}
\textbf{hyperparameter} & \textbf{TREC} & \textbf{Movie review} & \textbf{SST-2}\\
\hline
\noalign{\hrule height0.8pt}
Optimizer & Adam & Adam & Adam \\
Learning rate & 0.001 & 0.001 & 0.001 \\
Training epochs & 10 & 10 & 10 \\
$\alpha$ for $L_{KL}$ & 0.75 & 0.75 & 1 \\
Batch size & 50 & 50 & 50 \\
 \hline
 \noalign{\hrule height0.8pt}
 \end{tabular}
\caption{The hyperparameters of the experiment} 
\label{table:ap.lstm-hyper}
\end{table}

\section{Appendix D: Robustness Analysis} \label{sec:ap.robustness}

\noindent We evaluate the robustness of TrustAL by varying the hyperparameters that can affect the performance. The experiment is conducted on TREC with three AL algorithms we used across this paper.
The performance is the average performance of each stage. The relative performance difference from each stand-alone baseline is also presented in parenthesis where ``+" indicates performance gain and ``-" indicates performance degradation. The experimental results of TrustAL-NC are presented in Table (5-7) and TrustAL-MC are presented in Table (8-10).

\vspace{1mm}

\vspace{1mm}
\noindent\textbf{The Size of the Budget $\mathbf{k}$} The number of candidates for model selection in TrustAL-NC depends on the size of the budget $k$. We compare the performance of TrustAL-NC from low-budget to high-budget on three datasets. As a result, TrustAL-NC outperforms baseline regardless of the size of the budget $k$ in most cases. 

\vspace{1mm}
\noindent\textbf{The Size of the Validation Set}
We conduct an experiment to validate whether TrustAL-NC is robust on the size of development set. The experiment on TREC dataset shows that TrustAL-NC shows robust performance even with half-sized\footnote{Compared to the original setting} development set, resulting only 0.008 point decrease in accuracy on average over 3 AL strategies. We believe that these results relieve some concern about infeasibility of keeping development set in data scarce situations.




\vspace{1mm}
\noindent\textbf{Preference Weight for $\mathbf{L_{KD}}$} To analyze the effect of KD, we report the performance varying preference weights from 0.3 to 20. As a result, larger preference weight $\alpha$ mostly results in lower MCI. However, excessive preference weight seems result lower accuracy.

\begin{table}[h]\small
\setlength\extrarowheight{3pt}
\centering
 \begin{tabular}{c l c c c c}
 \hline
 \noalign{\hrule height 0.8pt}
 \multirow{2}{*}{} & &\multicolumn{2}{c}{\textbf{Stable}} &\multicolumn{2}{c}{\textbf{Saturated}}\\\cline{3-6}
  \multirow{2}{*}{} & &\textbf{acc.} & \textbf{MCI}  & \textbf{acc.} & \textbf{MCI}\\
\hline
\noalign{\hrule height0.8pt}
\multirow{2}{*}{A}    & baseline  & 0.759  & 159.6 & 0.844 & 70.0 \\\cline{2-6}
                         & TrustAL-NC  & 0.764  & 154.7  & 0.848 & 65.6 \\\cline{1-6}
\multirow{2}{*}{B} & baseline   & 0.757 & 136.9 & 0.846 & 64.9 \\\cline{2-6}
                         & TrustAL-NC & 0.758   & 118.0 & 0.840 & 60.0 \\\cline{1-6}
\multirow{2}{*}{C}   & baseline   & 0.769 & 123.9 & 0.851 & 60.0 \\\cline{2-6}
                         & TrustAL-NC & 0.769   & 114.2 & 0.844 & 59.0 \\\cline{1-6}
 \hline
 \noalign{\hrule height0.8pt}
 \end{tabular}
\caption{Average performance of TrustAL-NC using BERT with (A) CONF (B) CORESET and (C) BADGE on SST-2.} 
\label{table:sensitivity-bert}
\end{table}

\section{Appendix E: TrustAL-NC using BERT} \label{sec:ap.bert} 
To validate TrustAL-NC on different architecture, we report the performance of BERT\cite{devlin2019bert}.
We use Hugging Face's implementation\footnote{https://github.com/huggingface/transformers} of the \texttt{BERT-small} for efficiency.
We use the AdamW optimizer~\cite{loshchilov2018decoupled} with learning rate of 2e-5 with batch size 32, a max number of tokens of 128, epoch 3, adam epsilon 1e-8, gradients clipping to 5, distillation preference weight rate 1.5. TrustAL-NC using BERT outperforms the baseline regardless of the AL methods as it is shown in Table~\ref{table:sensitivity-bert}.

\section{Appendix F: Experimental Results} \label{sec:ap.experiment} 
Additional experimental results on Movie Review and SST-2 are presented in Figure 7 and Figure 8.

\section{Appendix G: Time Complexity of TrustAL-NC} \label{sec:ap.time} 
Here, we compare TrustAL-NC and query-by-committee \cite{seung1992query} in terms of time complexity to clarify the difference between two confusing concepts. Both utilize multiple models. However we claim TrustAL is (a) orthogonally
applicable to any AL strategies including query-by-committee
and (b) more efficient than query-by-committee
inference/training.
Regarding the inference time, TrustAL uses a single
model to evaluate each sample in $\calU$. TrustAL only requires
extra inference time to obtain pseudo labels which
can be optimized by simple caching trick to store the predictions
of training samples and make predictions only on non-overlapping
training examples which are newly acquired.
We found this technique to be especially efficient considering
the overlapping teacher selection tendency presented in
Figure 4. Also, regarding training time, TrustAL only trains
a student model and keeps the model fixed after a training
step. Query-by-committee usually trains
multiple committee models after each acquisition step which
is computationally expensive.

\begin{figure*}[t!]
     \centering
     \begin{subfigure}[b]{0.31\textwidth}
         \centering
         \includegraphics[width=\textwidth]{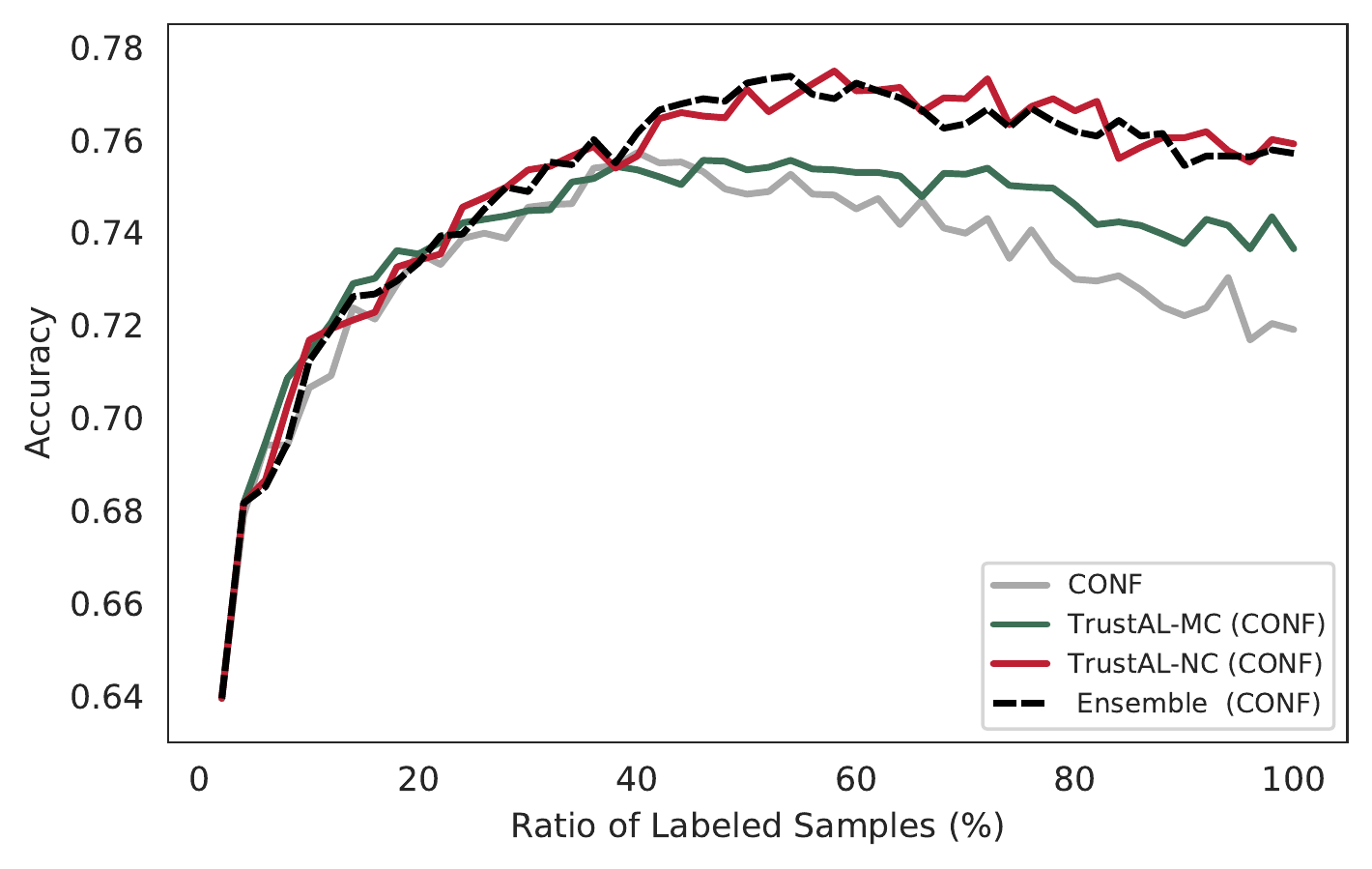}
         \caption{CONF}
     \end{subfigure}
     \hfill
     \begin{subfigure}[b]{0.31\textwidth}
         \centering
         \includegraphics[width=\textwidth]{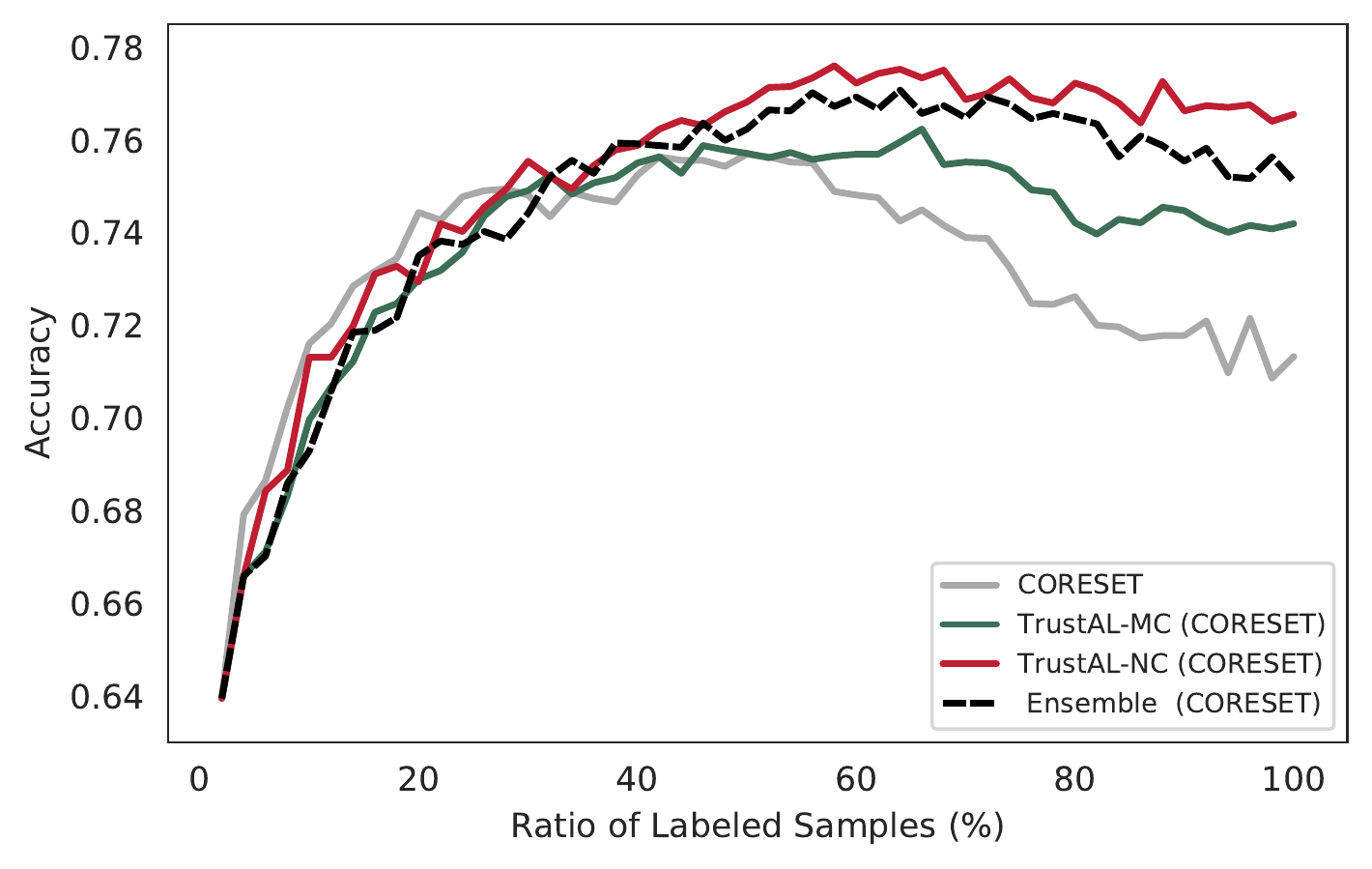}
         \caption{CORESET}
     \end{subfigure}
     \hfill
     \begin{subfigure}[b]{0.31\textwidth}
         \centering
         \includegraphics[width=\textwidth]{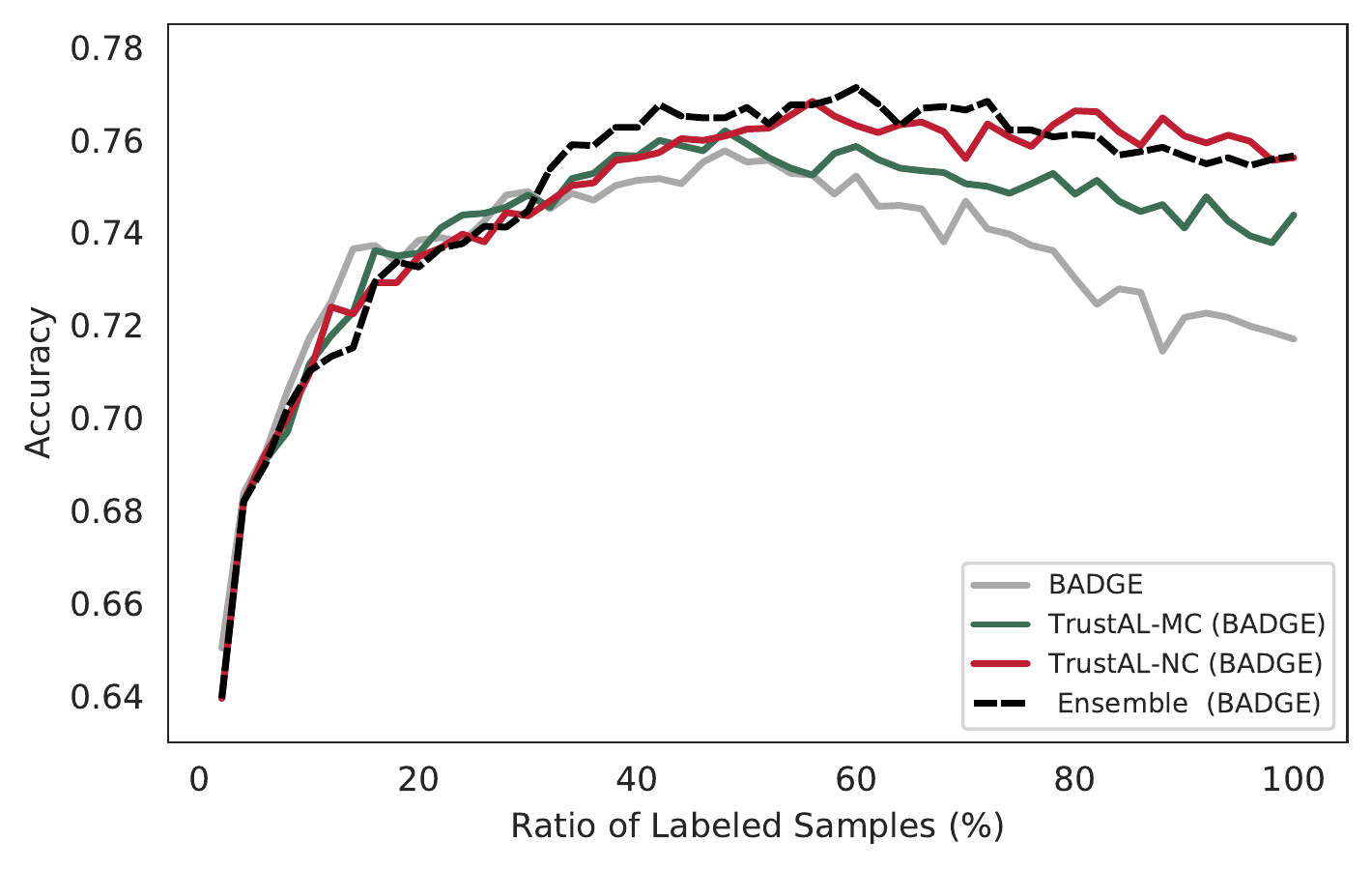}
         \caption{BADGE}
     \end{subfigure}
     \begin{subfigure}[b]{0.31\textwidth}
         \centering
         \includegraphics[width=\textwidth]{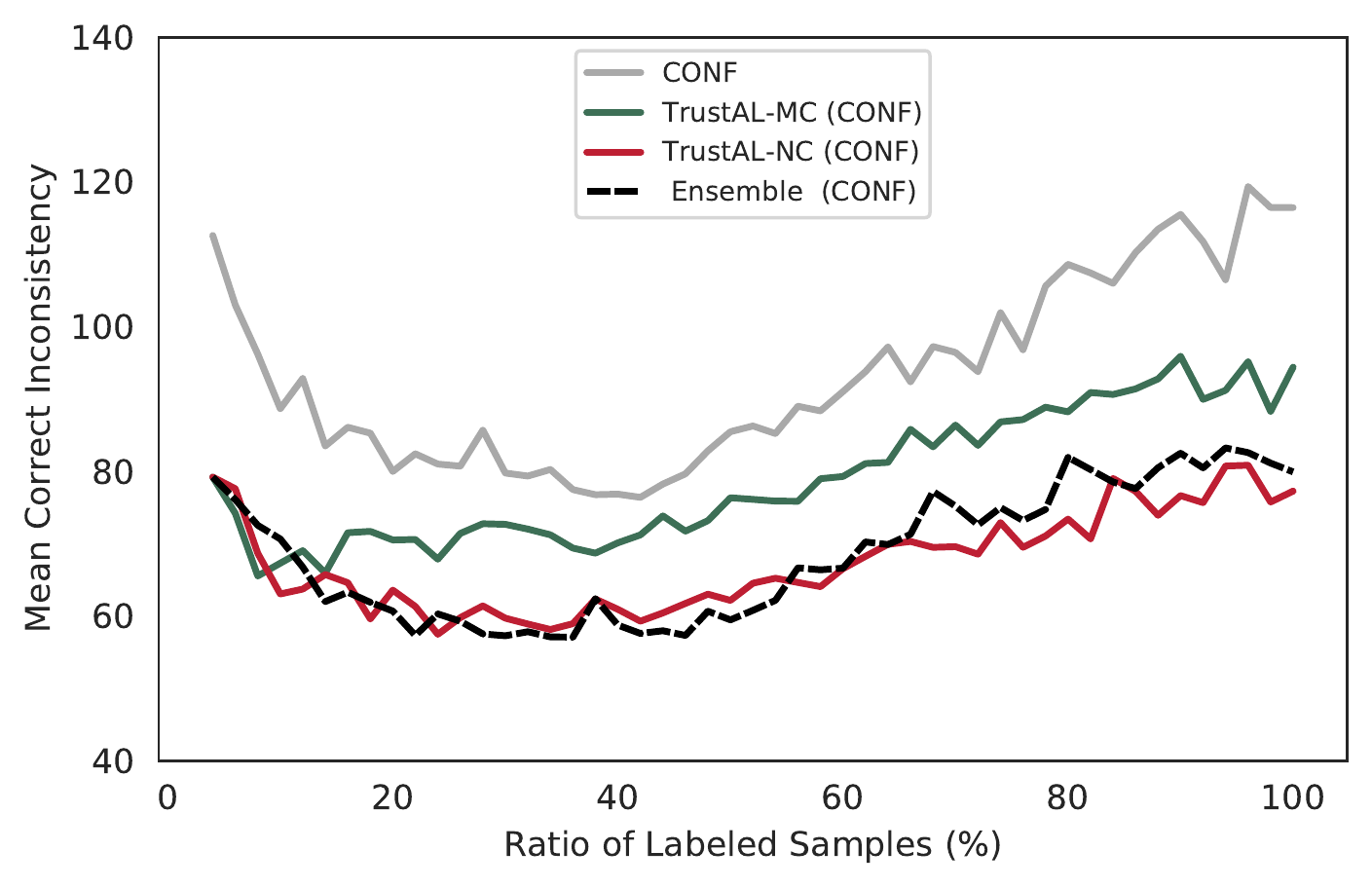}
         \caption{CONF}
     \end{subfigure}
     \hfill
     \begin{subfigure}[b]{0.31\textwidth}
         \centering
         \includegraphics[width=\textwidth]{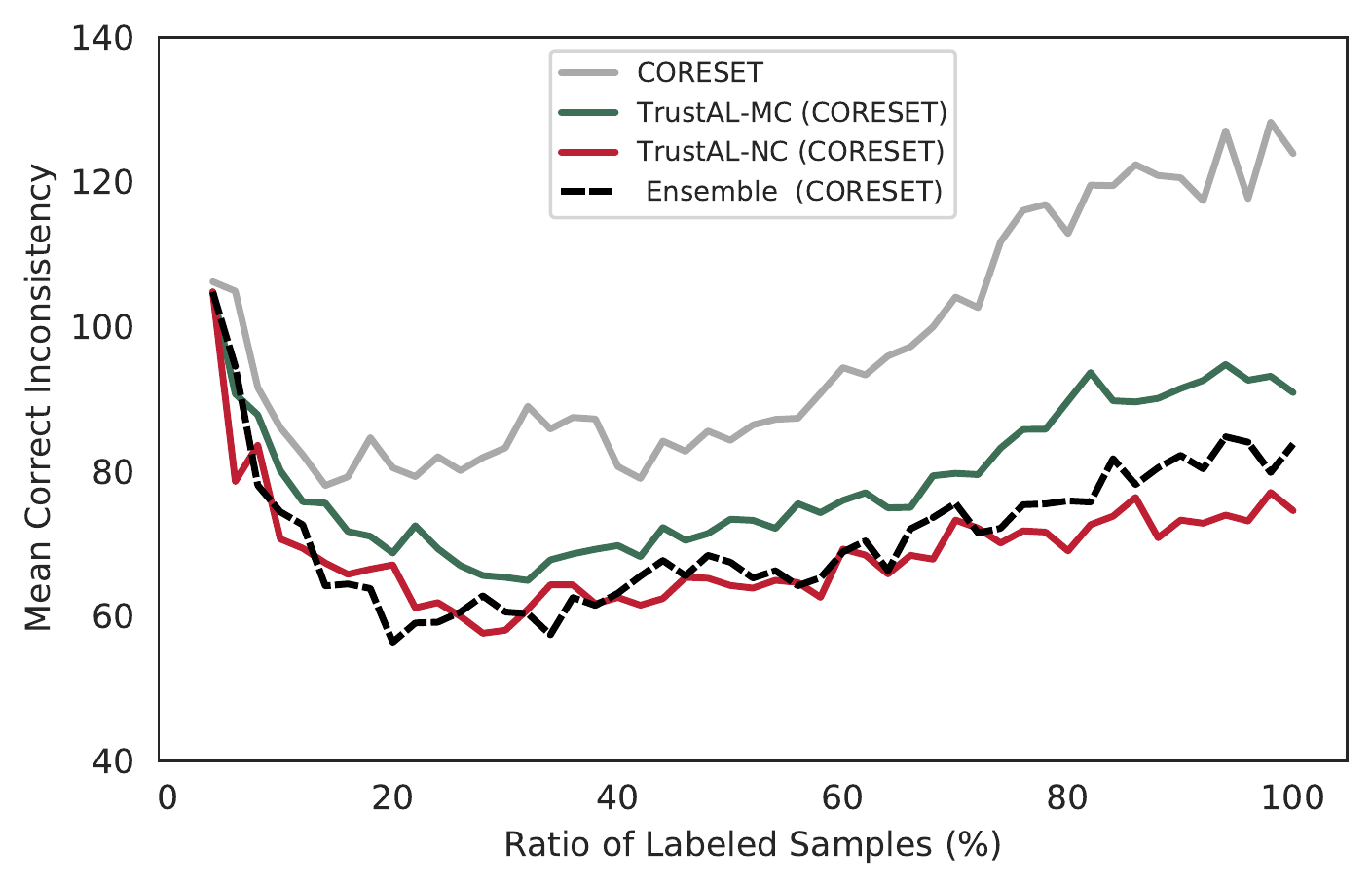}
         \caption{CORESET}
     \end{subfigure}
     \hfill
     \begin{subfigure}[b]{0.31\textwidth}
         \centering
         \includegraphics[width=\textwidth]{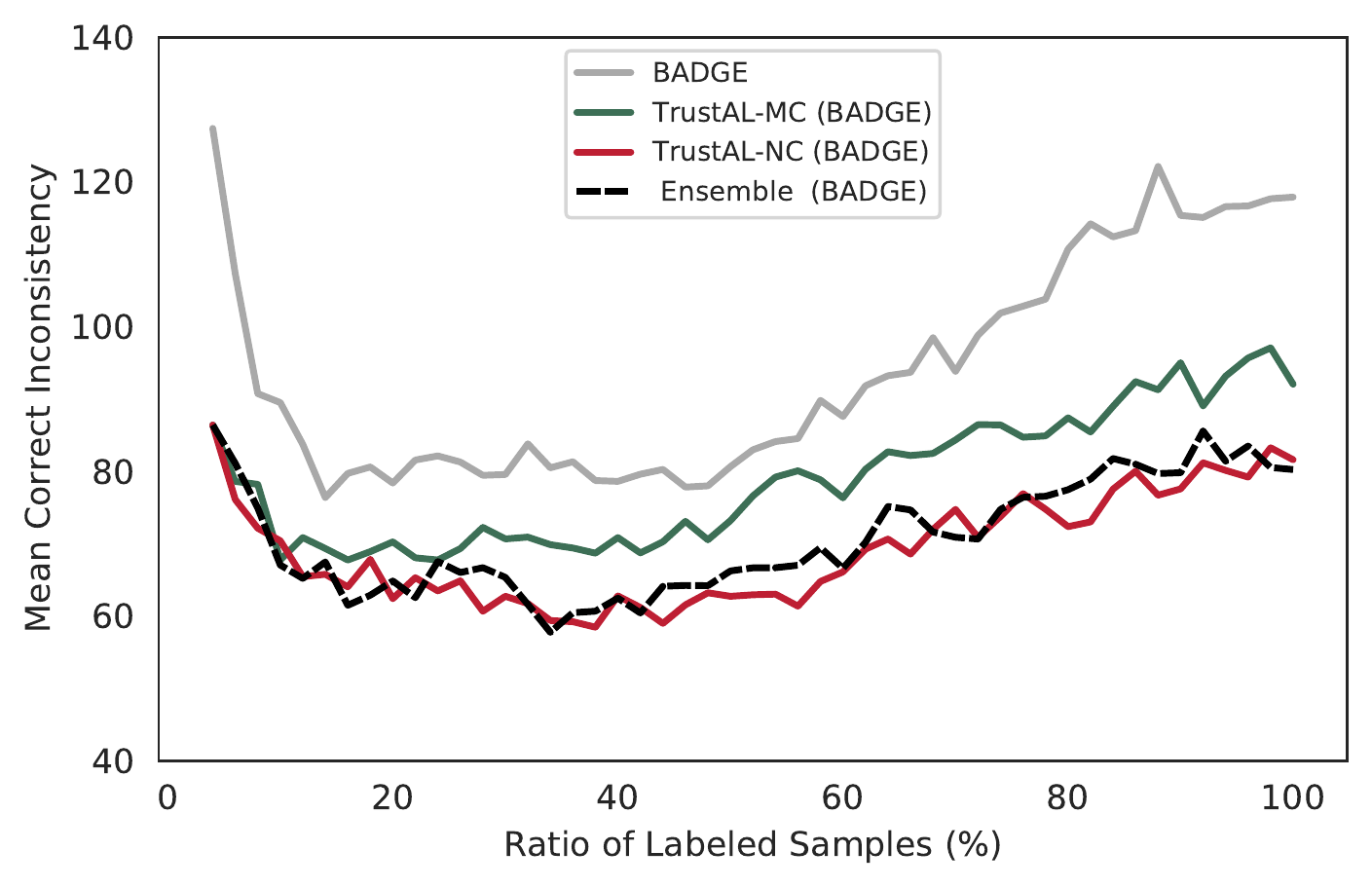}
         \caption{BADGE}
     \end{subfigure}
    \caption{Accuracy versus the ratio of labeled samples (a\text{-}c) and MCI versus the ratio of labeled samples (d\text{-}f) on Movie review test dataset.}
    \label{fig:rq1-moviereview}
\end{figure*}

\begin{figure*}[t!]
     \centering
     \begin{subfigure}[b]{0.31\textwidth}
         \centering
         \includegraphics[width=\textwidth]{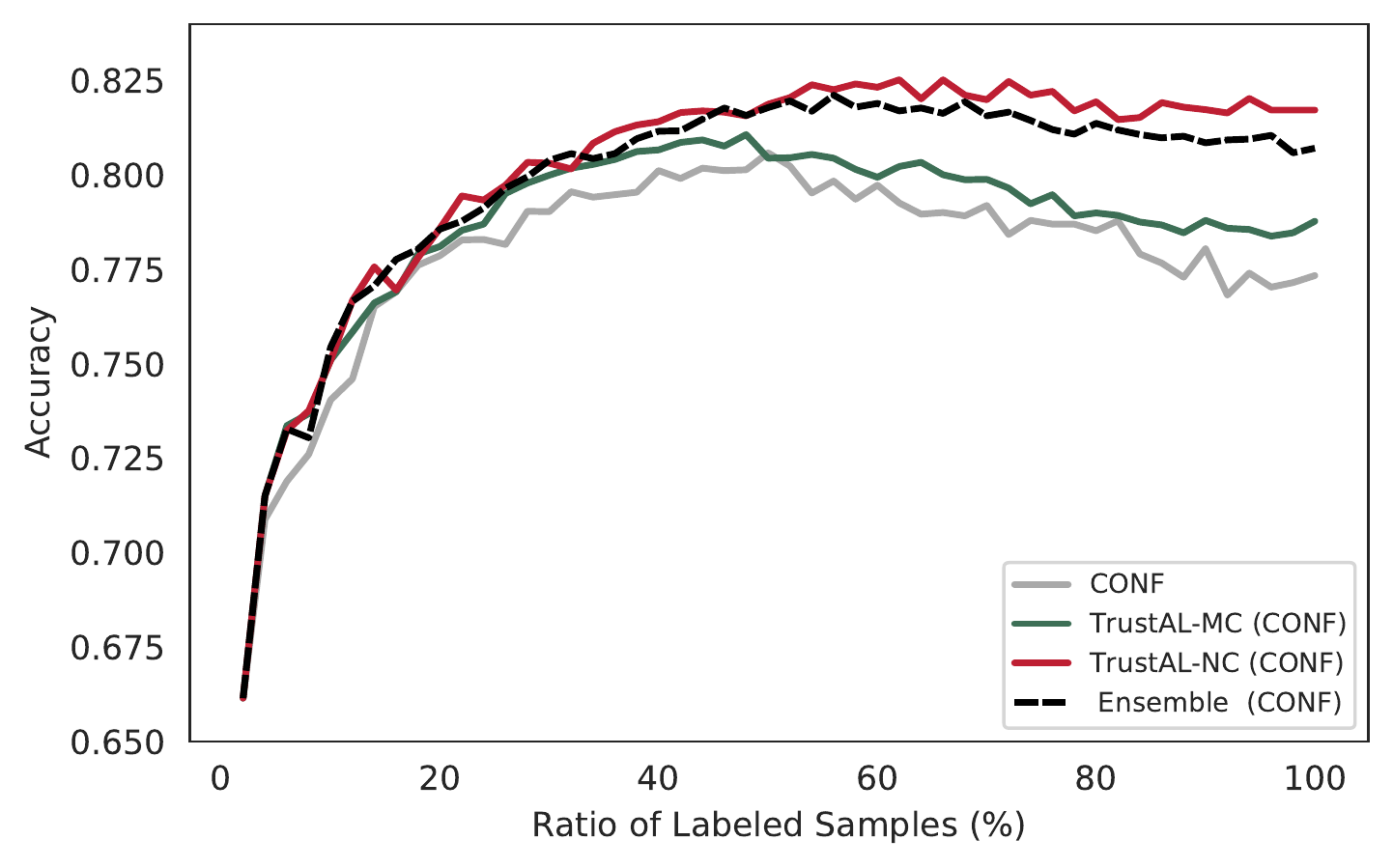}
         \caption{CONF}
     \end{subfigure}
     \hfill
     \begin{subfigure}[b]{0.31\textwidth}
         \centering
         \includegraphics[width=\textwidth]{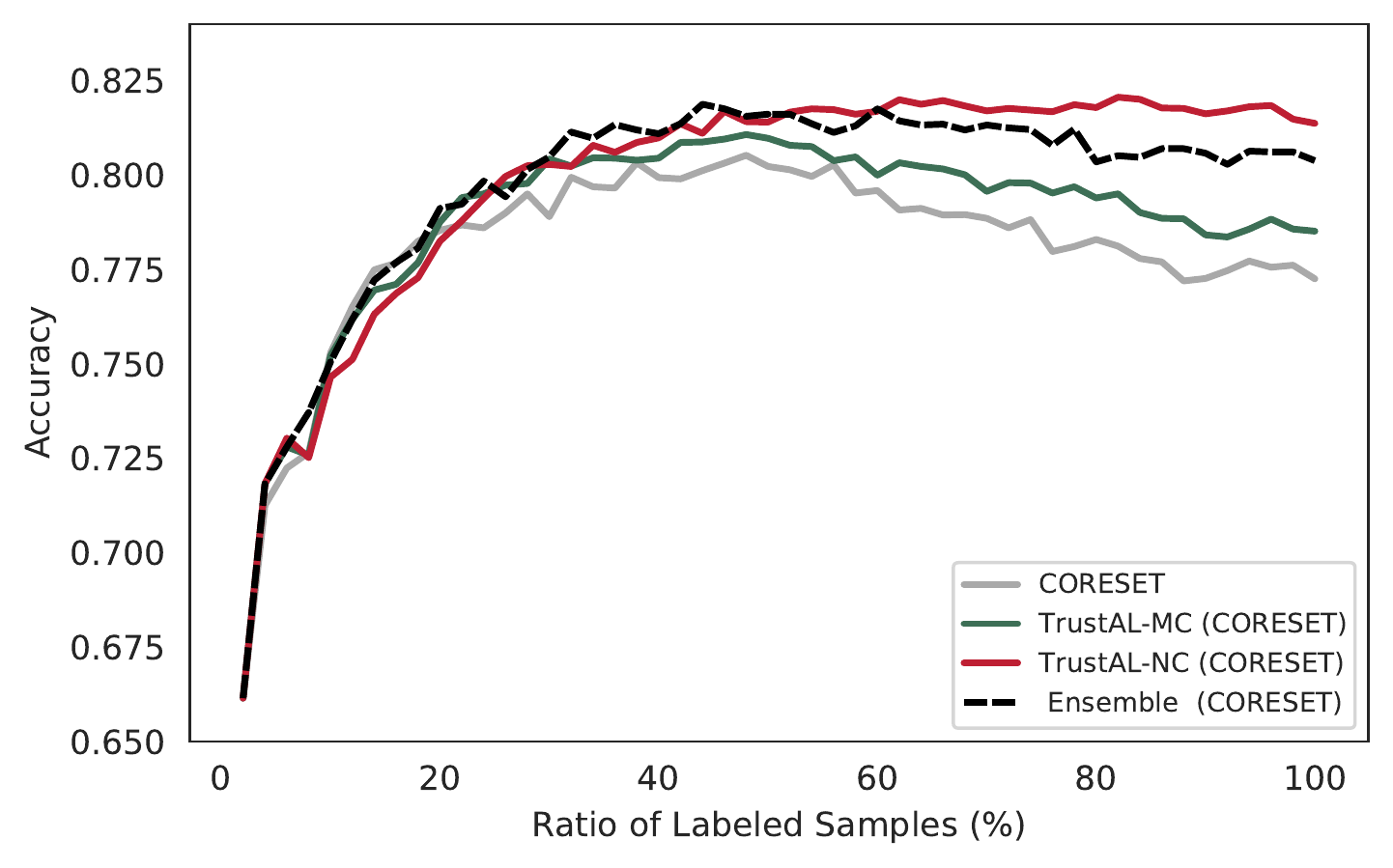}
         \caption{CORESET}
     \end{subfigure}
     \hfill
     \begin{subfigure}[b]{0.31\textwidth}
         \centering
         \includegraphics[width=\textwidth]{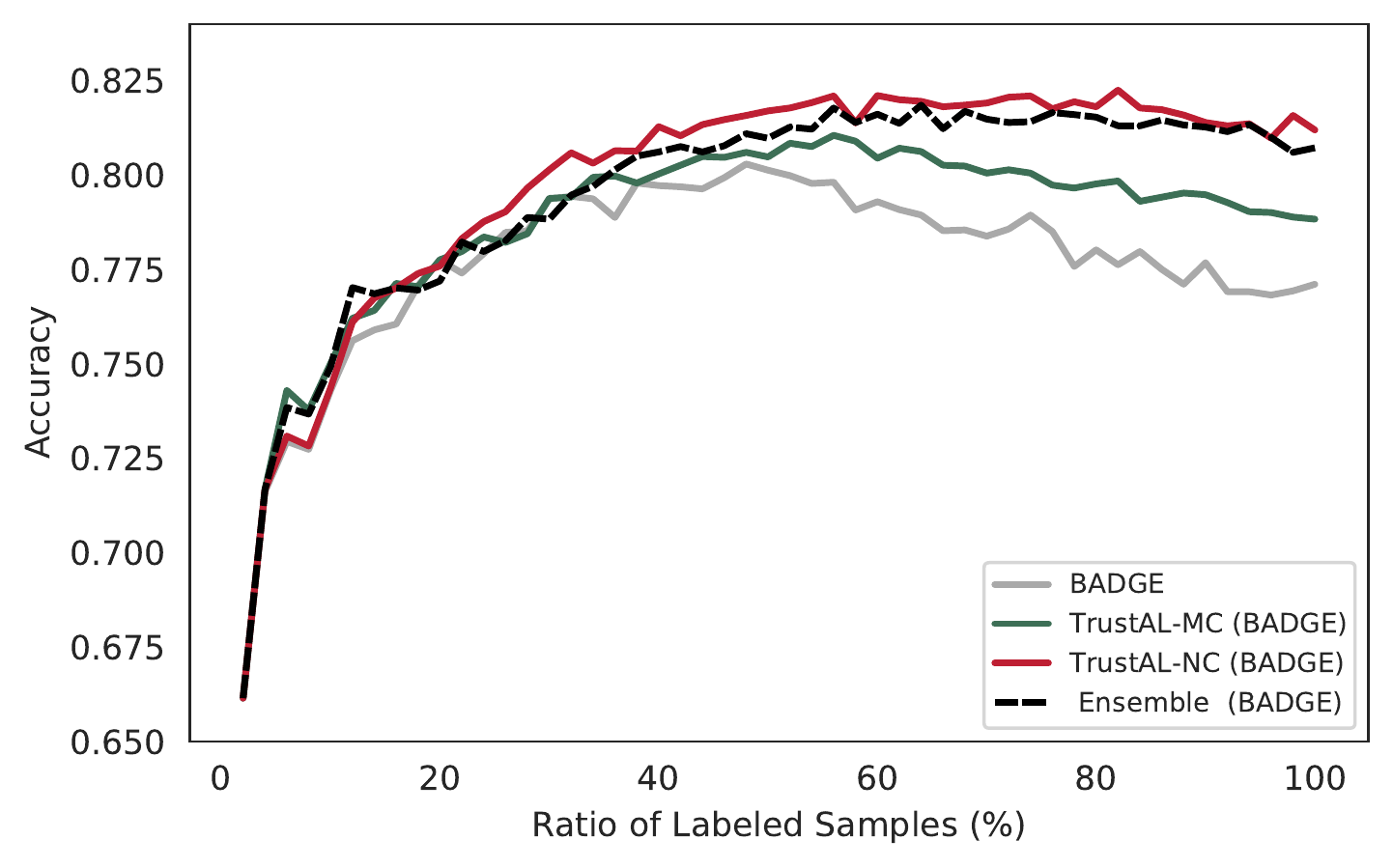}
         \caption{BADGE}
     \end{subfigure}
     \begin{subfigure}[b]{0.31\textwidth}
         \centering
         \includegraphics[width=\textwidth]{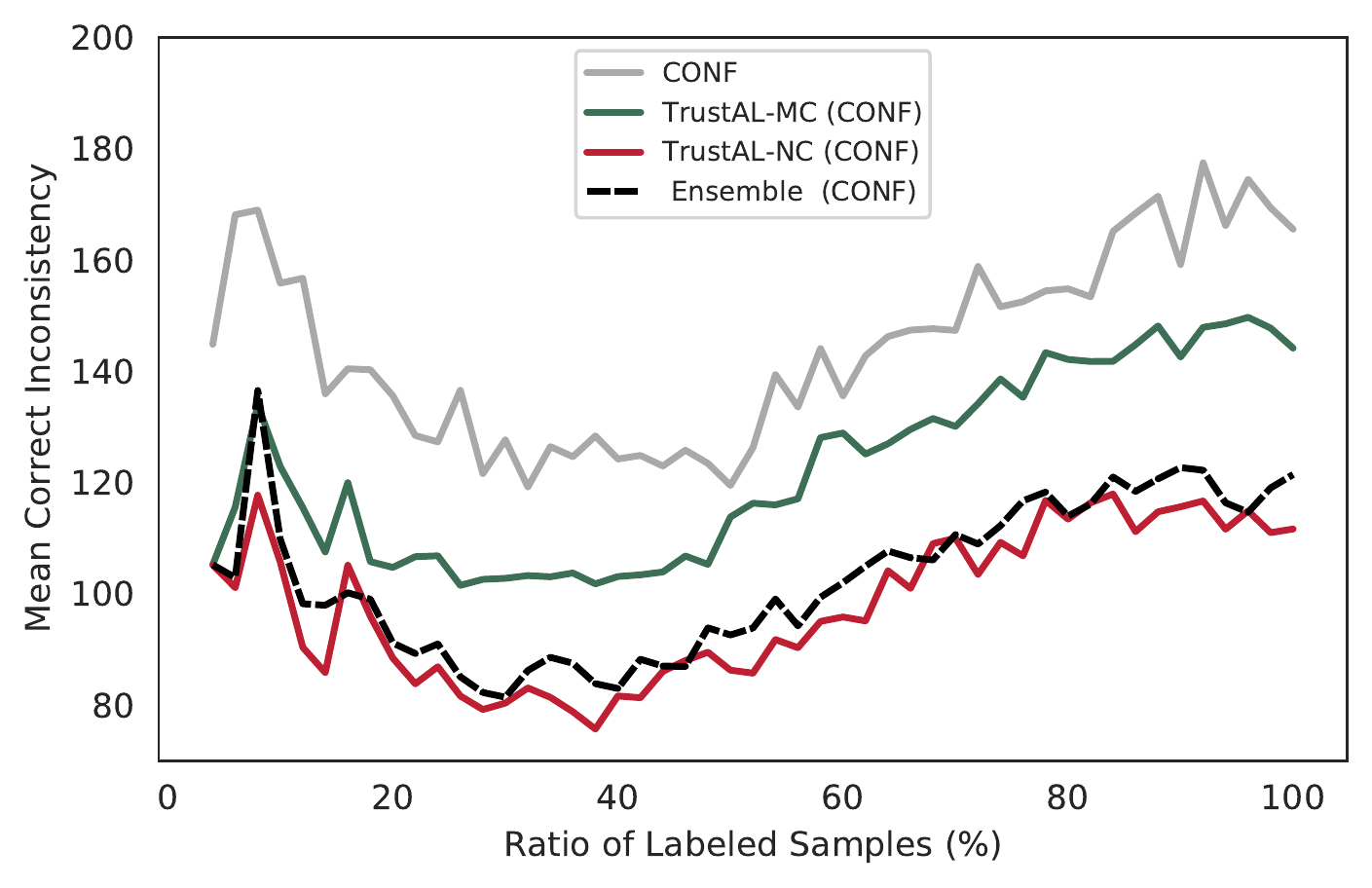}
         \caption{CONF}
     \end{subfigure}
     \hfill
     \begin{subfigure}[b]{0.31\textwidth}
         \centering
         \includegraphics[width=\textwidth]{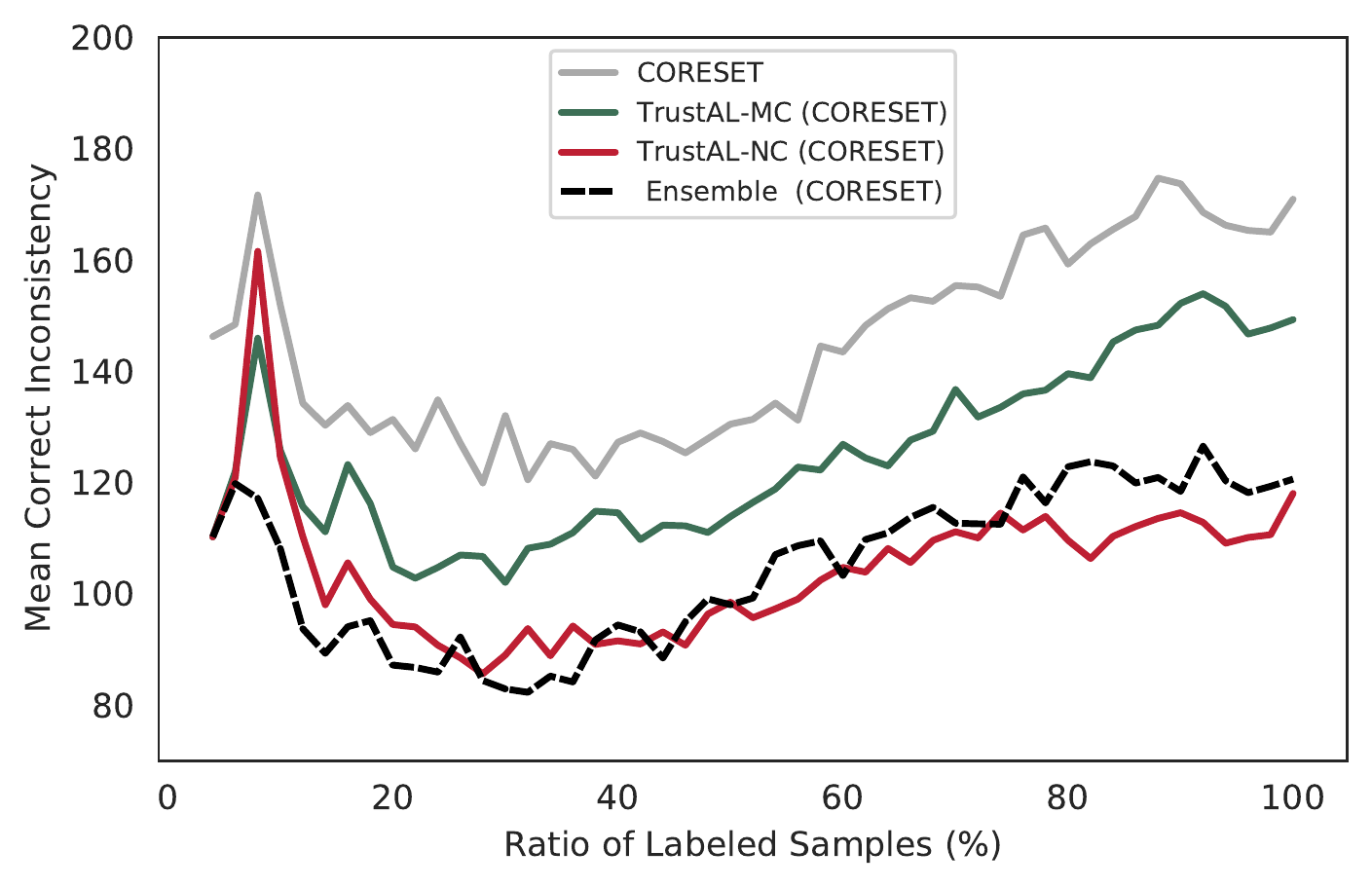}
         \caption{CORESET}
     \end{subfigure}
     \hfill
     \begin{subfigure}[b]{0.31\textwidth}
         \centering
         \includegraphics[width=\textwidth]{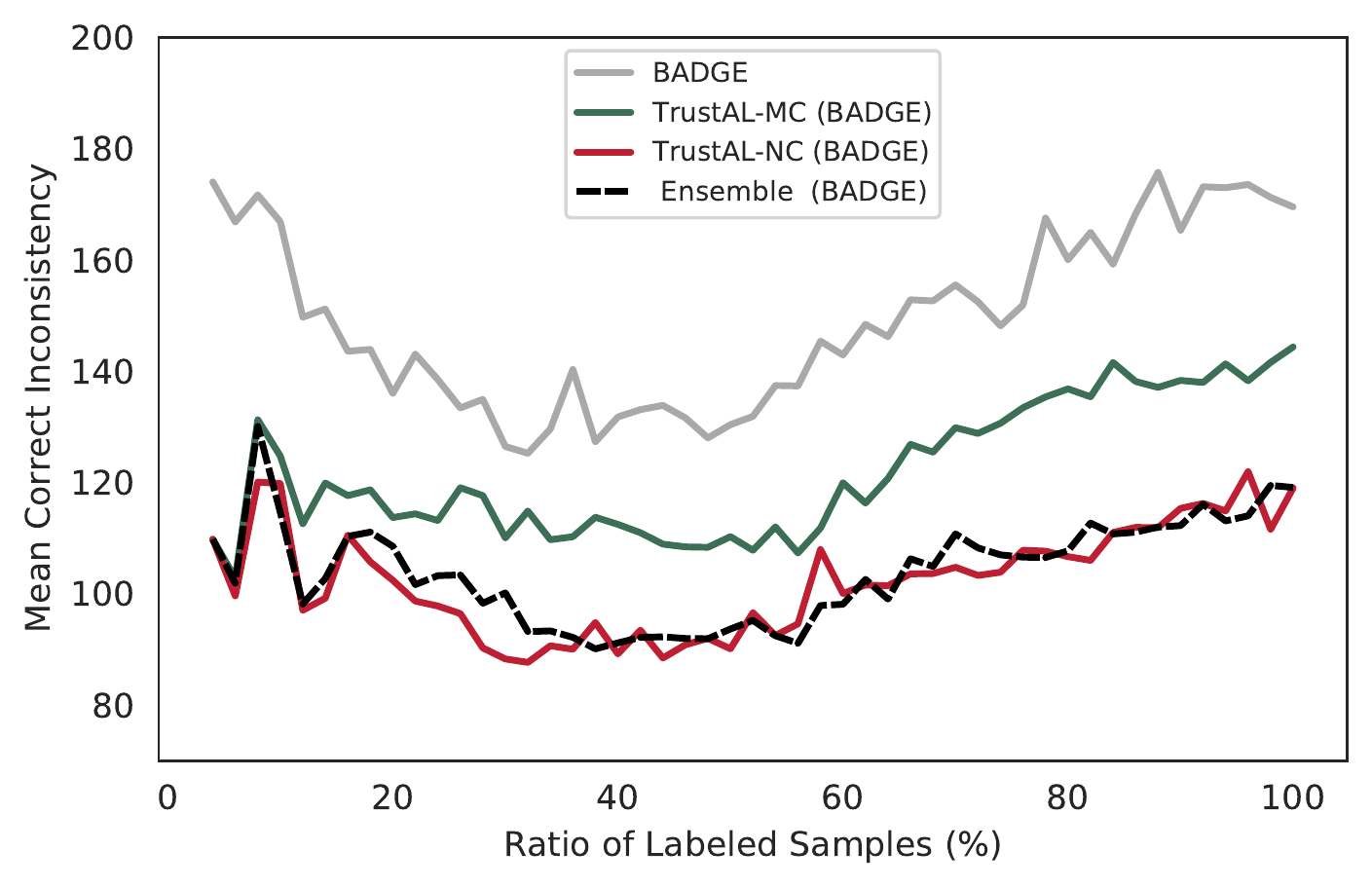}
         \caption{BADGE}
     \end{subfigure}
    \caption{Accuracy versus the ratio of labeled samples (a\text{-}c) and MCI versus the ratio of labeled samples (d\text{-}f) on SST-2 test dataset.}
    \label{fig:ap.performance-sst}
\end{figure*}


\begin{table*}[t]\small
\setlength\extrarowheight{3pt}
\centering
 \begin{tabular}{c c c c c c}
 \hline
 \noalign{\hrule height 0.8pt}
 \multirow{2}{*}{} & &\multicolumn{2}{c}{\textbf{Stable}} &\multicolumn{2}{c}{\textbf{Saturated}}\\\cline{3-6}
                  & &\textbf{acc.} & \textbf{CI}  & \textbf{acc.} & \textbf{CI}\\
\hline
\noalign{\hrule height0.8pt}
\multirow{3}{*}{Ratio of $D_{dev}$} & 0.5  & 0.792(+0.022) & 25.02({-5.80}) & 0.898({+0.042}) & 13.90({-11.12}) \\\cline{2-6}
                              & 1.0* & 0.788({+0.011}) & 17.94({-5.83}) & 0.896({+0.062}) & 13.64({-14.43}) \\\cline{2-6}
                              & 2.0 & 0.784({+0.018}) & 22.08({-4.49}) & 0.876({+0.044}) & 17.30({-10.76}) \\\cline{1-6}
\multirow{3}{*}{Size of budget $k$} & 0.02* & 0.788({+0.011}) & 17.94({-5.83}) & 0.896({+0.062}) & 13.64({-14.43}) \\\cline{2-6}
                              & 0.04 & 0.802({+0.016}) & 15.99({-3.38}) & 0.906({+0.041}) & 11.02({-11.52}) \\\cline{2-6}
                              & 0.1 & 0.825({-0.002}) & 8.17({-7.48}) & 0.896({+0.019}) & 11.06({-8.11}) \\\cline{1-6}
\multirow{5}{*}{$\alpha$ for $L_{KL}$} & 0.3 & 0.786({+0.009}) & 23.86({+0.09}) & 0.891({+0.058}) & 16.57({-11.50}) \\\cline{2-6}
                              & 0.75* & 0.788({+0.011}) & 17.94({-5.83}) & 0.896({+0.062}) & 13.64({-14.43}) \\\cline{2-6}
                              & 1.5 & 0.802({+0.026}) & 16.94({-6.83}) & 0.895({+0.062}) & 11.56({-16.50}) \\\cline{2-6}
                              & 10 & 0.810({+0.033}) & 10.62({-13.15}) & 0.892({+0.059}) & 9.44({-18.62}) \\\cline{2-6}
                              & 20 & 0.793({+0.016}) & 10.47({-13.30}) & 0.874({+0.041}) & 8.57({-19.49}) \\\cline{1-6}
 \hline
 \noalign{\hrule height0.8pt}
 \end{tabular}
\caption{Performance of TrustAL-NC with CONF on TREC for robustness analysis. The hyperparameter of the experiment on RQ1 is marked with *. } 
\label{table:ap.sensitivity-nc-conf}
\end{table*}

\begin{table*}[t]\small
\setlength\extrarowheight{3pt}
\centering
 \begin{tabular}{c c c c c c}
 \hline
 \noalign{\hrule height 0.8pt}
 \multirow{2}{*}{} & &\multicolumn{2}{c}{\textbf{Stable}} &\multicolumn{2}{c}{\textbf{Saturated}}\\\cline{3-6}
                  & &\textbf{acc.} & \textbf{CI}  & \textbf{acc.} & \textbf{CI}\\
\hline
\noalign{\hrule height0.8pt}
\multirow{3}{*}{Ratio of $D_{dev}$} & 0.5  & 0.785({+0.012}) & 25.08({-2.98}) & 0.908({+0.062})  & 11.18({-16.03}) \\\cline{2-6}
                              & 1.0* & 0.808({+0.035}) & 18.06({-7.24}) & 0.884({+0.038}) & 14.89({-13.34}) \\\cline{2-6}
                              & 2.0 & 0.782({+0.019}) & 23.50({-3.83}) & 0.885({+0.061}) & 15.48({-18.35}) \\\cline{1-6}
\multirow{3}{*}{Size of budget $k$} & 0.02* & 0.808({+0.035}) & 18.06({-7.24}) & 0.884({+0.038}) & 14.89({-13.34}) \\\cline{2-6}
                              & 0.04 & 0.798({+0.014}) & 13.21({-6.65}) & 0.896({+0.026}) & 11.72({-9.76}) \\\cline{2-6}
                              & 0.1 & 0.823({+0.005}) & 10.85({-4.38}) & 0.895({+0.018}) & 10.21({-6.53}) \\\cline{1-6}
\multirow{5}{*}{$\alpha$ for $L_{KL}$} & 0.3 & 0.793({+0.019}) & 18.51({-6.79}) & 0.895({+0.048}) & 13.96({-14.27}) \\\cline{2-6}
                              & 0.75* & 0.808({+0.035}) & 18.06({-7.24}) & 0.884({+0.038}) & 14.89({-13.34}) \\\cline{2-6}
                              & 1.5 & 0.811({+0.037}) & 14.33({-10.96}) & 0.895({+0.049}) & 10.74({-17.50}) \\\cline{2-6}
                              & 10 & 0.819({+0.046}) & 12.54({-12.75}) & 0.899({+0.052}) & 8.84({-19.39}) \\\cline{2-6}
                              & 20 & 0.803({+0.029}) & 9.94({-15.36}) & 0.891({+0.044}) & 8.21({-20.02}) \\\cline{1-6}
 \hline
 \noalign{\hrule height0.8pt}
 \end{tabular}
\caption{Performance of TrustAL-NC with CORESET on TREC for robustness analysis. The hyperparameter of the experiment on RQ1 is marked with *. } 
\label{table:ap.sensitivity-nc-coreset}
\end{table*}

\begin{table*}[t]\small
\setlength\extrarowheight{3pt}
\centering
 \begin{tabular}{c c c c c c}
 \hline
 \noalign{\hrule height 0.8pt}
 \multirow{2}{*}{} & &\multicolumn{2}{c}{\textbf{Stable Learning Stage}} &\multicolumn{2}{c}{\textbf{Saturated Learning Stage}}\\\cline{3-6}
                  & &\textbf{accuracy} & \textbf{MCI} & \textbf{accuracy} & \textbf{MCI} \\
\hline
\noalign{\hrule height0.8pt}
\multirow{3}{*}{Ratio of $D_{dev}$} & 0.5  & 0.792({+0.037}) & 22.47({-9.53})  & 0.897({+0.063}) & 13.40({-14.90}) \\\cline{2-6}
                              & 1.0* & 0.796({+0.017}) & 17.13({-4.61}) & 0.896({+0.048}) & 14.08({-11.49}) \\\cline{2-6}
                              & 2.0 & 0.791({+0.011}) & 19.12({-5.13}) & 0.889({+0.053}) & 15.96({-14.32}) \\\cline{1-6}
                              
\multirow{3}{*}{Size of budget $k$} & 0.02* & 0.796({+0.017}) & 17.13({-4.61}) & 0.896({+0.048}) & 14.08({-11.49}) \\\cline{2-6}
                              & 0.04 & 0.787({-0.002}) & 16.78({-1.19}) & 0.911({+0.040}) & 10.84({-10.77}) \\\cline{2-6}
                              & 0.1 & 0.816({+0.010}) & 10.98({-6.23}) & 0.897({-0.004}) & 13.21({-2.00}) \\\cline{1-6}
\multirow{5}{*}{$\alpha$ for $L_{KL}$} & 0.3 & 0.794({+0.015}) & 19.28({-2.47}) & 0.884({+0.036}) & 17.53({-8.04}) \\\cline{2-6}
                              & 0.75* & 0.796({+0.017}) & 17.13({-4.61}) & 0.896({+0.048}) & 14.08({-11.49}) \\\cline{2-6}
                              & 1.5 & 0.792({+0.013}) & 16.79({-4.96}) & 0.904({+0.056}) & 10.73({-14.84}) \\\cline{2-6}
                              & 10 & 0.797({+0.018}) & 13.81({-7.94}) & 0.891({+0.043}) & 10.66({-14.91}) \\\cline{2-6}
                              & 20 & 0.788({+0.010}) & 11.51({-10.23}) & 0.887({+0.039}) & 10.33({-15.24}) \\\cline{1-6}
 \hline
 \noalign{\hrule height0.8pt}
 \end{tabular}
\caption{Performance of TrustAL-NC with BADGE on TREC for robustness analysis. The hyperparameter of the experiment on RQ1 is marked with *. } 
\label{table:sensitivity-dev}
\end{table*}

\begin{table*}[t]\small
\setlength\extrarowheight{3pt}
\centering
 \begin{tabular}{c c c c c c}
 \hline
 \noalign{\hrule height 0.8pt}
 \multirow{2}{*}{} & &\multicolumn{2}{c}{\textbf{Stable}} &\multicolumn{2}{c}{\textbf{Saturated}}\\\cline{3-6}
                  & &\textbf{acc.} & \textbf{CI}  & \textbf{acc.} & \textbf{CI}\\
\hline
\noalign{\hrule height0.8pt}
                              
\multirow{3}{*}{Size of budget $k$} & 0.02* & 0.774({-0.003}) & 23.55({-0.22}) & 0.861({+0.028}) & 20.20({-7.86}) \\\cline{2-6}
                              & 0.04 & 0.797({+0.011}) & 19.98({+0.62}) & 0.883({+0.018}) & 16.51({-6.03}) \\\cline{2-6}
                              & 0.1 & 0.820({-0.007}) & 9.06({-6.58}) & 0.900({+0.024}) & 8.47({-10.69}) \\\cline{1-6}
                              
\multirow{5}{*}{$\alpha$ for $L_{KL}$} & 0.3 & 0.760({-0.016}) & 26.52({+2.75}) & 0.847({+0.014}) & 23.10({-4.97}) \\\cline{2-6}
                              & 0.75* & 0.774({-0.003}) & 23.55({-0.22}) & 0.861({+0.028}) & 20.20({-7.86}) \\\cline{2-6}
                              & 1.5 & 0.786({+0.009}) & 20.42({-3.35}) & 0.869({+0.035}) & 19.49({-8.57}) \\\cline{2-6}
                              & 10 & 0.799({+0.022}) & 16.45({-7.32}) & 0.879({+0.046}) & 15.41({-12.65}) \\\cline{2-6}
                              & 20 & 0.803({+0.026}) & 14.56({-9.21}) & 0.882({+0.048}) & 12.89({-15.17}) \\\cline{1-6}
 \hline
 \noalign{\hrule height0.8pt}
 \end{tabular}
\caption{Performance of TrustAL-MC with CONF on TREC for robustness analysis. The hyperparameter of the experiment on RQ1 is marked with *. Ratio of $D_{dev}$ is skipped for TrustAL-MC.} 
\label{table:ap.sensitivity-mc-conf}
\end{table*}

\begin{table*}[t]\small
\setlength\extrarowheight{3pt}
\centering
 \begin{tabular}{c c c c c c}
 \hline
 \noalign{\hrule height 0.8pt}
 \multirow{2}{*}{} & &\multicolumn{2}{c}{\textbf{Stable}} &\multicolumn{2}{c}{\textbf{Saturated}}\\\cline{3-6}
                  & &\textbf{acc.} & \textbf{CI}  & \textbf{acc.} & \textbf{CI}\\
\hline
\noalign{\hrule height0.8pt}
                              
\multirow{3}{*}{Size of budget $k$} & 0.02* & 0.791({+0.018}) & 21.45({-3.84}) & 0.853({+0.007}) & 22.92({-5.31}) \\\cline{2-6}
                              & 0.04 & 0.794({+0.010}) & 17.07({-2.79}) & 0.875({+0.005}) & 19.11({-2.37}) \\\cline{2-6}
                              & 0.1 & 0.820({+0.002}) & 12.23({-3.0}) & 0.900({+0.022}) & 9.96({-6.78}) \\\cline{1-6}
                              
\multirow{5}{*}{$\alpha$ for $L_{KL}$} & 0.3 & 0.776({+0.002}) & 21.62({-3.67}) & 0.839({-0.007}) & 27.78({-0.45}) \\\cline{2-6}
                              & 0.75* & 0.791({+0.018}) & 21.45({-3.84}) & 0.853({+0.007}) & 22.92({-5.31}) \\\cline{2-6}
                              & 1.5 & 0.795({+0.021}) & 20.08({-5.21}) & 0.852({+0.005}) & 22.97({-5.26}) \\\cline{2-6}
                              & 10 & 0.806({+0.033}) & 16.82({-8.48}) & 0.875({+0.029}) & 18.08({-10.15}) \\\cline{2-6}
                              & 20 & 0.811({+0.037}) & 13.66({-11.64}) & 0.890({+0.043}) & 13.01({-15.22}) \\\cline{1-6}
 \hline
 \noalign{\hrule height0.8pt}
 \end{tabular}
\caption{Performance of TrustAL-MC with CORESET on TREC for robustness analysis. The hyperparameter of the experiment on RQ1 is marked with *. Ratio of $D_{dev}$ is skipped for TrustAL-MC.} 
\label{table:ap.sensitivity-mc-coreset}
\end{table*}

\begin{table*}[t]\small
\setlength\extrarowheight{3pt}
\centering
 \begin{tabular}{c c c c c c}
 \hline
 \noalign{\hrule height 0.8pt}
 \multirow{2}{*}{} & &\multicolumn{2}{c}{\textbf{Stable}} &\multicolumn{2}{c}{\textbf{Saturated}}\\\cline{3-6}
                  & &\textbf{acc.} & \textbf{CI}  & \textbf{acc.} & \textbf{CI}\\
\hline
\noalign{\hrule height0.8pt}
                              
\multirow{3}{*}{Size of budget $k$} & 0.02* & 0.779({+0.001}) & 23.68({+1.93}) & 0.862({+0.014}) & 20.02({-5.55}) \\\cline{2-6}
                              & 0.04 & 0.787({-0.001}) & 17.76({-0.21}) & 0.89({+0.019}) & 16.06({-5.55}) \\\cline{2-6}
                              & 0.1 & 0.808({+0.002}) & 17.04({-0.17}) & 0.901({+0.000}) & 11.30({-3.91}) \\\cline{1-6}
                              
\multirow{5}{*}{$\alpha$ for $L_{KL}$} & 0.3 & 0.774({-0.004}) & 24.82({+3.07}) & 0.855({+0.007}) & 22.40({-3.17}) \\\cline{2-6}
                              & 0.75* & 0.779({+0.001}) & 23.68({+1.93}) & 0.862({+0.014}) & 20.02({-5.55}) \\\cline{2-6}
                              & 1.5 & 0.787({+0.008}) & 18.84({-2.90}) & 0.874({+0.026}) & 21.27({-4.30}) \\\cline{2-6}
                              & 10 & 0.818({+0.040}) & 16.01({-5.74}) & 0.871({+0.023}) & 22.27({-3.29}) \\\cline{2-6}
                              & 20 & 0.789({+0.011}) & 15.84({-5.91}) & 0.878({+0.030}) & 13.26({-12.31}) \\\cline{1-6}
 \hline
 \noalign{\hrule height0.8pt}
 \end{tabular}
\caption{Performance of TrustAL-MC with BADGE on TREC for robustness analysis. The hyperparameter of the experiment on RQ1 is marked with *. Ratio of $D_{dev}$ is skipped for TrustAL-MC.} 
\label{table:ap.sensitivity-mc-badge}
\end{table*}
\end{document}